\pdfoutput=1

\documentclass[conference]{IEEEtran}
\usepackage{amsmath}
\usepackage{graphicx}
\usepackage{caption}
\usepackage{subcaption}
\usepackage{array}
\usepackage{verbatim,hyperref}
\pdfoutput=1
\begin{document}

\title{Tomographic reconstruction using \\ global statistical priors}

\author{\IEEEauthorblockN{Preeti Gopal,$^{1,3}$ Ritwick Chaudhry,$^{2}$ Sharat Chandran,$^{2}$ Imants Svalbe,$^{3}$ Ajit Rajwade$^{2}$}
\IEEEauthorblockA{$^{1}$Dept. of CSE, IIT Bombay, and IITB-Monash Research Academy\\
$^{2}$Dept. of CSE, IIT Bombay, India\\
$^{3}$School of Physics and Astronomy, Monash University, Australia\\
\{preetig, ritwick, sharat, ajitvr\}@cse.iitb.ac.in, imants.svalbe@monash.edu}}

\maketitle

\begin{abstract}
Recent research in tomographic reconstruction is motivated by the need
to efficiently recover detailed anatomy from limited measurements. One
of the ways to compensate for the increasingly sparse sets of
measurements is to exploit the information from \emph{templates},
i.e., prior data available in the form of already reconstructed,
structurally similar images. Towards this,
previous work has exploited using a set of global and patch based
dictionary priors. In this paper, we propose a global prior to improve
both the speed and quality of tomographic reconstruction within a
Compressive Sensing framework.

We choose a set of potential representative 2D images referred to as
templates, to build an eigenspace; this is subsequently used to guide
the iterative reconstruction of a similar slice from sparse
acquisition data. Our experiments across a diverse range of datasets
show that reconstruction using an appropriate global prior, apart from
being faster, gives a much lower reconstruction error when compared to
the state of the art.
\end{abstract}
\IEEEpeerreviewmaketitle

\section{Introduction}
\label{sec:intro}
Improvements in tomographic reconstruction from sparse measurements
are being widely sought after, due to their benefit of aiding lower
exposure to the ionizing X-rays. However, the quality of
reconstruction is directly limited by the number of measurements
captured. When sampling is below the Nyquist limit, the data cannot be
uniquely reconstructed by direct inversion methods. Hence, given an
exceedingly smaller number of measurements, the reconstruction must
rely on some known information about the data for better
reconstruction fidelity. This information compensates for the
sub-Nyquist sampling of the data. The information could be, for
example, the sparsity of the data under some transform. Compressive
sensing (CS)~\cite{Donoho} routines exploit this information to
recover the sparse coefficients of the underlying data, with proven
guarantees for error bounds.

Our previous work~\cite{Preeti2016} partitioned a volume comprised of consecutive 2D slices into local groups of slices before performing the reconstruction of this
volume from a sparse set of projected views. The grouping of these slices was based on their structural similarity, as measured  through a comparison of the invariant moments of their projected views. Acquiring local knowledge of the object structure enabled significant improvement, as now the CS reconstruction could be performed on groups of similar slices that were able to be reconstructed much better than the individual, slice by slice CS reconstructions. In addition to this, recent research has also focussed on using known information that is \emph{specific} to the type of data being measured, for example, an existing similar data, also known as the \emph{prior}. Such a technique has been referred to as Prior Image Constrained Compressed Sensing (PICCS)~\cite{PICCS}. Here, the similarity between the prior and the reconstructed data is explicitly enforced in the iterative routines. This offers further reduction in sampling when compared to plain CS, for a desired reconstruction quality. 

However, in scenarios where the similarity of the prior and test data are not guaranteed, the application of PICCS will hurt the reconstruction. Moreover, selecting \emph{any one particular} optimal slice or volume as a prior from a huge amount of available data, can be challenging. To evade this shortcoming, a large amount of work has been done on the usage of \emph{a set} of slices or volumes as prior data~\cite{liu2016},~\cite{Xu2012},~\cite{song2014},~\cite{Li2016},~\cite{Lior2015}. The similarity of the test slice to any one of the templates is then more probable. One application scenario for using a prior is in the case of long-term treatments, in which a subject might undergo CT scan multiple times. The subsequent scans can then be taken with fewer projections by exploiting the information available from all or a few of the previous good quality scans. 

\textbf{Contributions:} In this work, we develop on the idea of using multiple templates by hypothesizing that the test slice may not be \emph{exactly} similar to any of the slices in the template set, but lies in the subspace spanned by the templates. In such a case, the test slice can be expressed as a linear combination of the principal components of the template slices. Hence, we form an eigenspace from a set of template slices and use this as our prior data. We have compared our results with plain CS, filtered back-projection and patch based dictionary methods and observed a substantial improvement with highly undersampled and noisy measurements. 

In Section~\ref{sec:lit}, we discuss the existing literature in
prior-based reconstructions. In Section~\ref{sec:our_method}, we
describe the proposed global-prior-based technique, followed by the
patch based dictionary technique in Section~\ref{sec:patch_based}. The
results of our experiments on five CT datasets are described in
Section~\ref{sec:experiments}, followed by some conclusions, and
future work in Section~\ref{sec:happy_ending}.

\section{Previous work}
\label{sec:lit}

The quality of image reconstruction depends on the type of recovery algorithm used and the way in which the prior is used. Among the various types of iterative recovery algorithms, Compressed Sensing (CS)~\cite{Donoho} is being widely used  in tomographic reconstruction due to its theoretical guarantee~\cite{candes} of powerful data recovery from noisy sub-Nyquist measurements.

Among the prior aided methods, a direct extension of the PICCS based method is the usage of multiple priors. This is demonstrated in~\cite{Luong_2016}, where the optimal weights to the set of priors is adaptively tuned after each iteration of reconstruction. Many methods have also been proposed recently to exploit local prior data. In~\cite{liu2016}, volume reconstruction from low dose CT was performed using a 3D patch based overcomplete dictionary built offline from a set of good quality scan volumes. The performance of this dictionary was shown to be superior to Simultaneous Iterative Reconstruction Technique (SIRT), 3D-TV and 2D slice-wise patch based dictionary. There has also been previous work based on the update of the 2D patch based dictionary during the reconstruction process, based on the intermediate reconstructions~\cite{Xu2012}. This is however computationally very expensive. In~\cite{song2014}, dual dictionaries have been built for 3D MRI volume reconstruction, one corresponding to low resolution volume and other for high resolution volume. In this way, the similarity between the volumes with different resolutions is exploited. Since this dual-dictionary is fixed and sensitive to parameters of reconstruction, an adaptive dual dictionary was built in~\cite{Li2016}. The adaptive version was performed on 2D patches. In ~\cite{Lior2015}, both the MRI sampling position and the weight given to the prior are varied in real-time based on the differences between a single template volume and the test volume. Instead of using a set of templates,~\cite{Zhan2016} had used one template image and its patches were classified into multiple classes based on their geometrical orientation. A separate dictionary was then learned on each class of patch. The intermediate reconstructions were used to update the template slice and the whole process was repeated. 

In contrast to all the previous work, we propose to use a more accurate and faster global Principal Component Analysis (PCA) based dictionary as part of the CS scheme.
   \begin{figure}
\centering
\begin{subfigure}[b]{0.95\linewidth}
        \includegraphics[width=\linewidth]{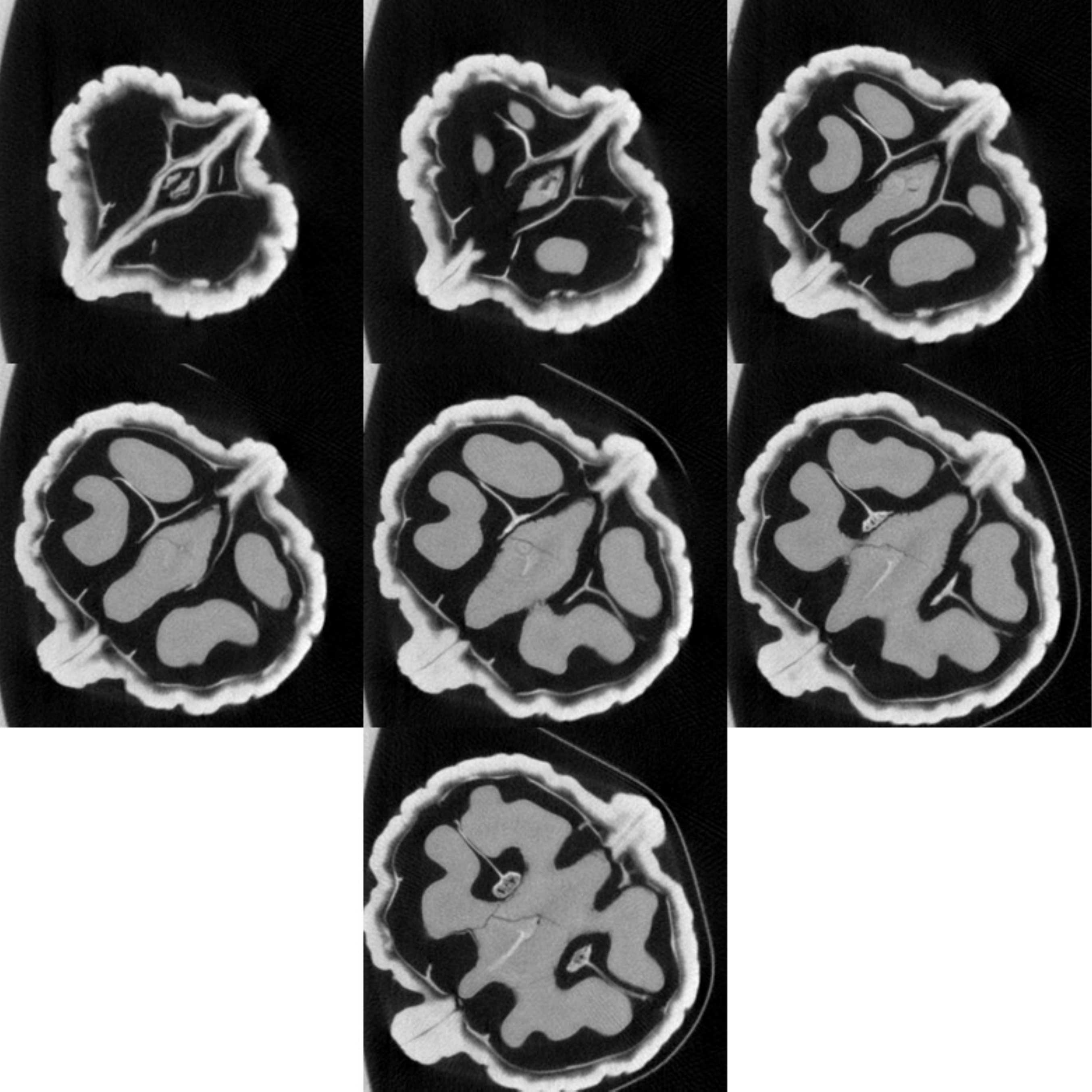}
\caption{}
\label{fig:walnut_templates}
\end{subfigure}
    \begin{subfigure}[b]{0.32\linewidth}
        \includegraphics[width=\textwidth]{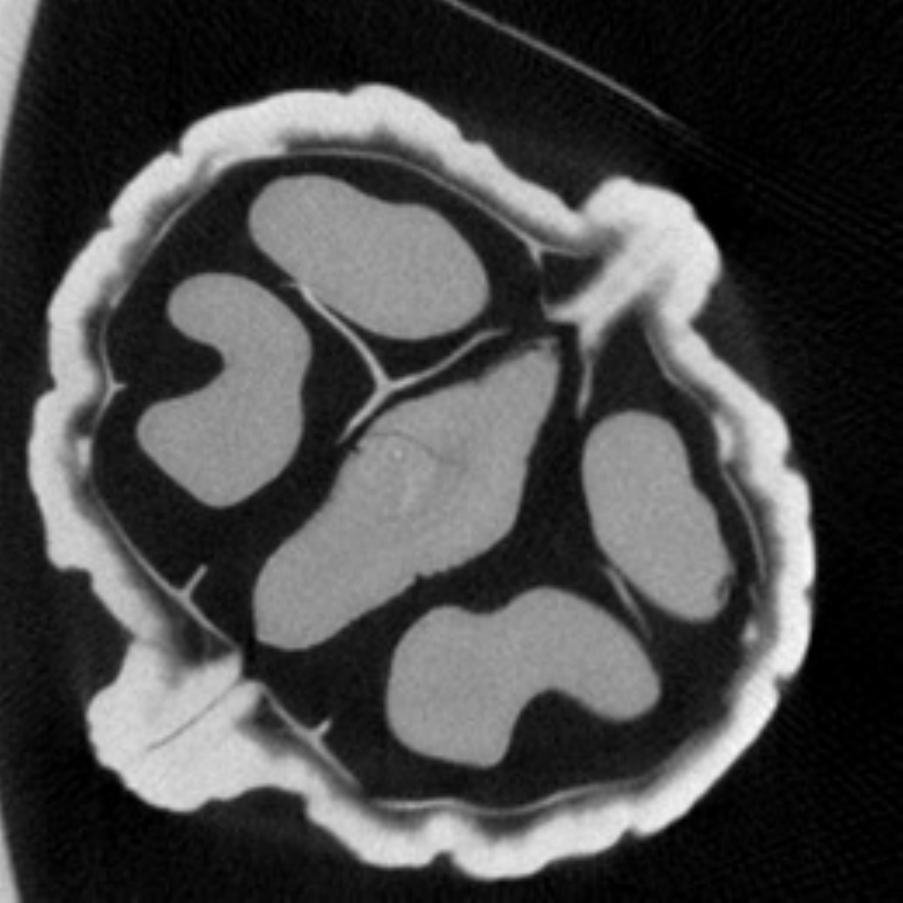}
        \caption{}
\label{fig:walnut_test}
     \end{subfigure}
        \caption{Data from walnut CT dataset~\cite{walnut}. Each slice is of size $260\times 260$. (a) Seven templates (slice nos. 70, 80, 90, 100, 110, 120 and 130 from the CT volume) that were used to build the prior. (b) The test data (slice no. 105 from the same volume). }

    \end{figure}

\begin{figure*}
\centering
    \begin{subfigure}[b]{0.7\linewidth}
\centering
        \includegraphics[width=0.9\textwidth]{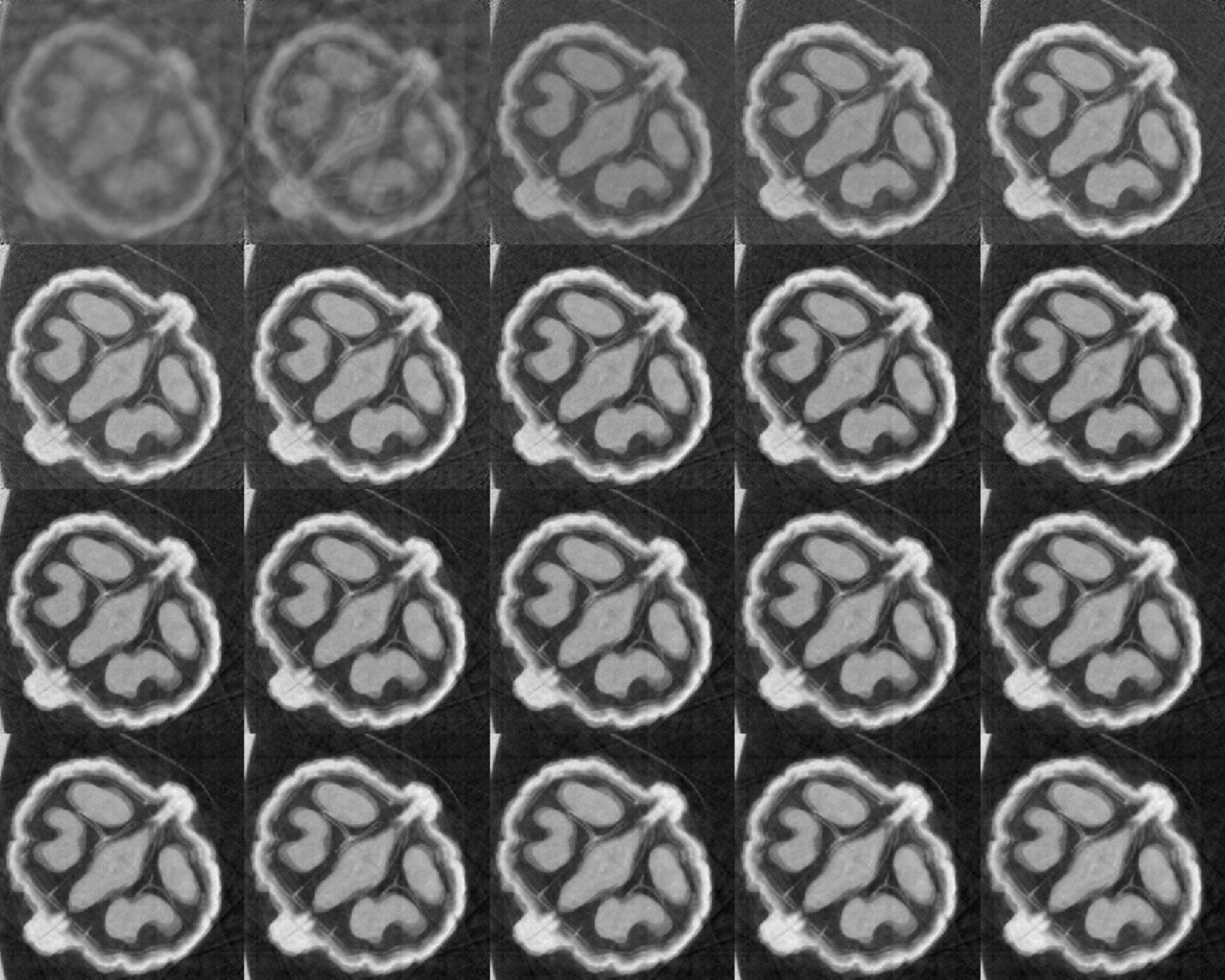}    
\caption{}   
     \end{subfigure}
\par\bigskip 
    \begin{subfigure}[b]{0.7\linewidth}
\centering
        \includegraphics[width=1\textwidth]{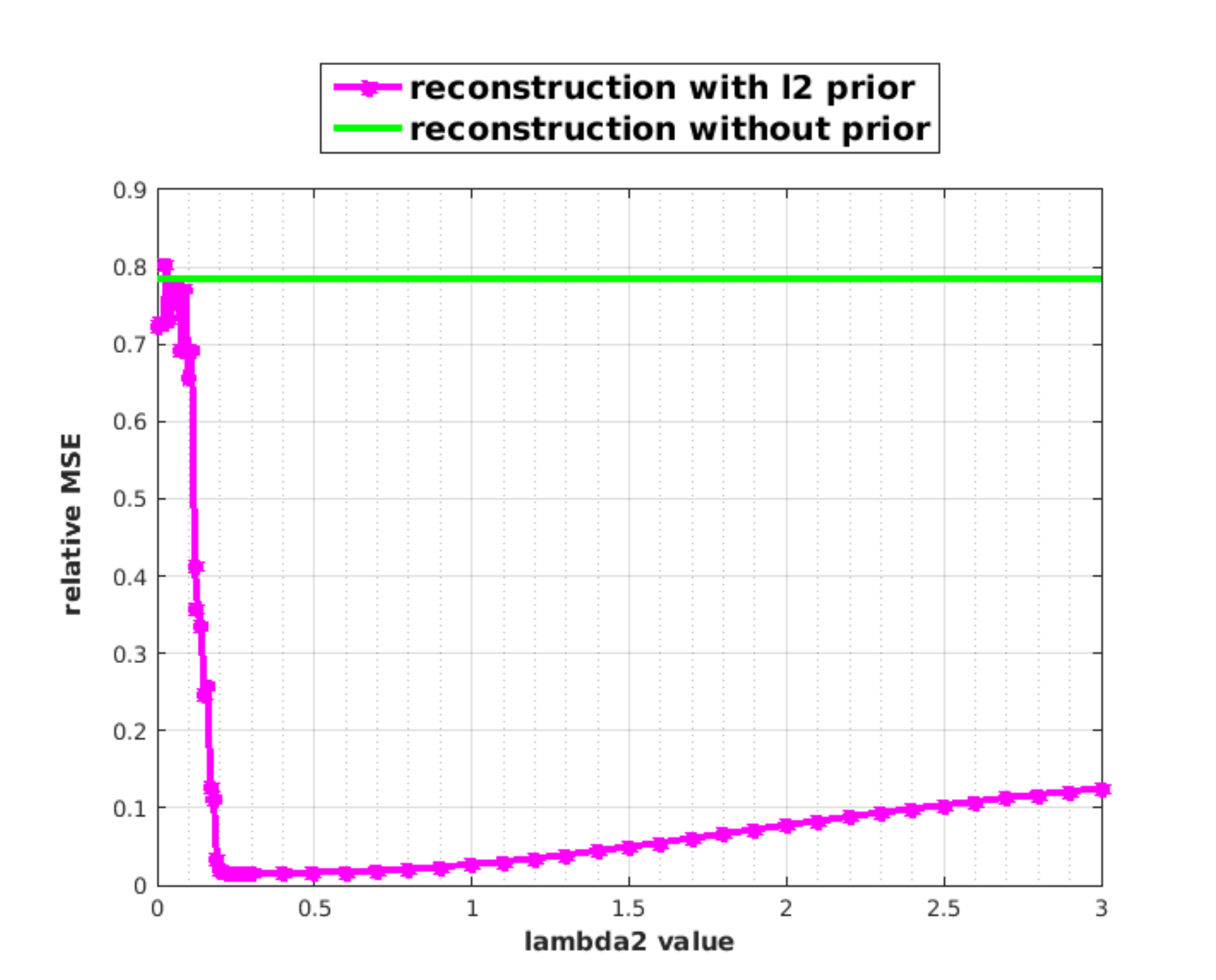}
     \caption{}
     \end{subfigure}
    \caption{Offline tuning for $\lambda_2$ for the use of global PCA
      based prior in the walnut CT dataset~\cite{walnut} to
      reconstruct the test slice shown in
      figure~\ref{fig:walnut_test}. (a) reconstruction for
      $\lambda_2$ varying from $0$ to $2$ ins-eps-converted-to.pdf of $0.1$ (b) 
      variation in relative mean squared error for different
      values of $\lambda_2$.}
\label{fig:tuning}
\end{figure*}
\section{Computing global statistical prior}
\label{sec:our_method}
 We begin with the assumption that the test slice can be expressed as a compact linear combination of the principal components extracted from a set of similar slices (called here as `templates'). Hence, we build the prior using PCA, which has been a widely used technique to represent data in terms of its significant modes or directions of variation. 

In our algorithm, we choose a set of representative 2D slices as templates. These templates must represent a wide range of anatomy in order to contribute to an eigenspace that can encompass a range of possible test slice anatomy. If these slices are from different volumes, then they must be first registered before computing the prior. The prior is built by computing the covariance matrix $C$ from the template set $\{t_i\}$, as shown in Eq.~\ref{eq:PCA}, The space spanned by the eigenvectors $\{V_i\}$ (eigenspace) of the covariance matrix is our global prior and is assumed to contain any test slice that is similar, but not necessarily identical to the templates.

The covariance matrix is given by:
 \begin{equation}
\begin{split}
  C  & = \frac{1}{N-1}\sum_{i=1}^N(t_i - \mu)(t_i - \mu)^{T}\\
\end{split}
\label{eq:PCA}
\end{equation}
where $\mu$ denotes the mean of all templates $\{t_i\}$ and $N$ denotes the number of templates chosen.  

Once the prior eigenspace is learned offline, the reconstruction is
performed iteratively by minimizing an objective function. This
minimization is done using the Compressive Sensing (CS) based
routines. Let $\Phi$ denote the Radon measurement matrix under a
chosen set of angles, $\theta$ denote the sparse 2D-DCT (Discrete
Cosine Transform) coefficients of the slice to be reconstructed, and
$\Psi$ denote the inverse DCT basis matrix. The DCT is commonly used
to sparsely represent naturally occurring images in the CS
framework. The objective function to be minimized in the plain CS
technique is given by:

\begin{equation}
E(\mathbf{\theta}) = ||\Phi\Psi\mathbf{\theta}- y||_2^2  + \lambda_1||\mathbf{\theta}||_1
\label{Eq:plain_CS}
\end{equation}
After inclusion of the PCA based prior, this objective function $E$ is given by :

\begin{equation}
E(\mathbf{\theta},\mathbf{\alpha}) = ||\Phi\Psi\mathbf{\theta}- y||_2^2  + \lambda_1||\mathbf{\theta}||_1+\lambda_2||\Psi\mathbf{\theta} - (\mu + \sum_{i}V_i\alpha_i)||_2^2
\label{Eq:main}
\end{equation}
where $V_i$ denotes the $i^{th}$ principal component or eigenvector
and $\alpha$ denotes the eigen coefficients of $\Psi\mathbf{\theta}$
(an estimate of the underlying test slice). We solve for $\theta$ and
$\alpha$ by alternately minimizing Eq.~\ref{Eq:primaryObj} by assuming
a fixed $\alpha$, and Eq.~\ref{Eq:alpha}, by using a fixed
$\mathbf{\theta}$. Eq.~\ref{Eq:primaryObj} is solved using the $l1\_ls$
solver~\cite{l1ls}, one of the popular solvers used in the literature.

\begin{equation}
E_{\mathbf{\alpha}}(\mathbf{\theta}) = ||\Phi\Psi\mathbf{\theta}- y||_2^2  + \lambda_1||\mathbf{\theta}||_1+\lambda_2||\Psi\mathbf{\theta} - (\mu + \sum_{i}V_i\alpha_i)||_2^2
\label{Eq:primaryObj}
\end{equation}
\begin{equation}
E_{\mathbf\theta}(\alpha) = ||\Psi\mathbf{\theta} - (\mu + \sum_{i}V_i\alpha_i)||_2^2
\label{Eq:alpha1}
\end{equation}
Eq.~\ref{Eq:alpha1} leads to the following closed form update for $\alpha$:
\begin{equation}
\mathbf{\mathbf{\alpha}} = \mathbf{V}^T(\Psi\mathbf{\theta} - \mu)
\label{Eq:alpha}
\end{equation}
An optimal value of $\lambda_2$ is empirically chosen \textit{a priori}, based on the reconstructions of one of the template slices. The variation in relative mse (relative mean squared error) for different weights of the template based prior, $\lambda_2$ is shown in figure~\ref{fig:tuning}.

\begin{figure}
\centering
    \begin{subfigure}[b]{0.24\linewidth}
        \includegraphics[width=\textwidth]{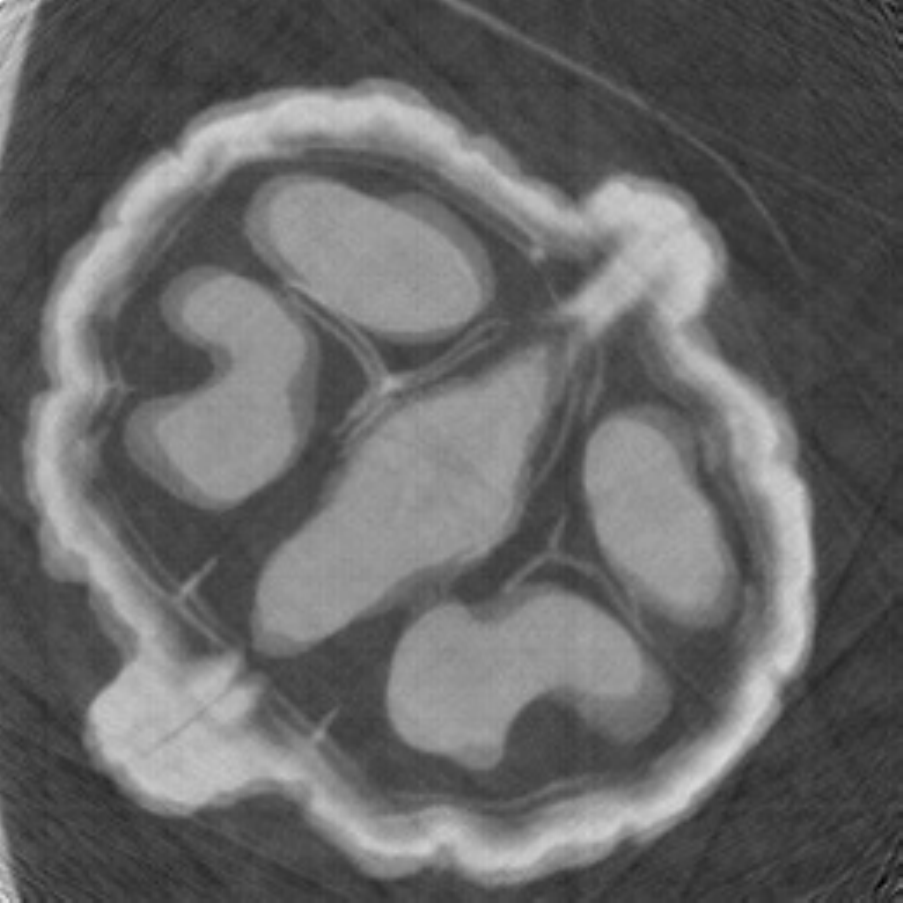}
        \caption{}
     \end{subfigure}
    \begin{subfigure}[b]{0.24\linewidth}
        \includegraphics[width=\textwidth]{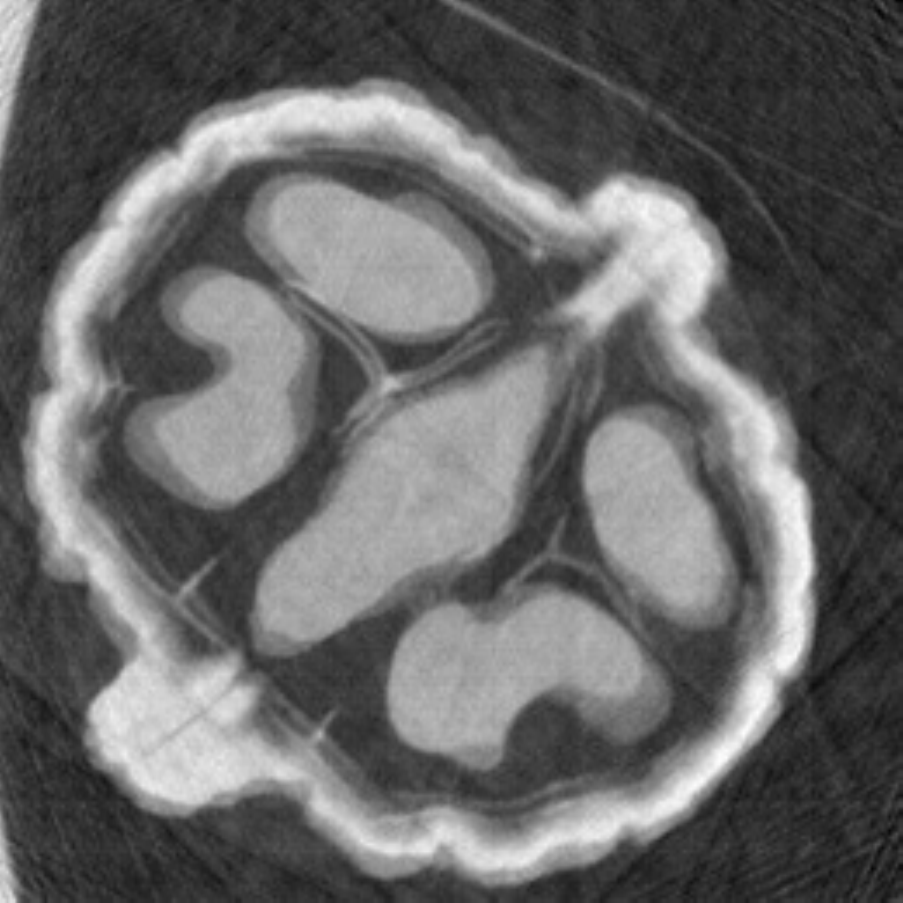}
        \caption{}
     \end{subfigure}
    \begin{subfigure}[b]{0.24\linewidth}
        \includegraphics[width=\textwidth]{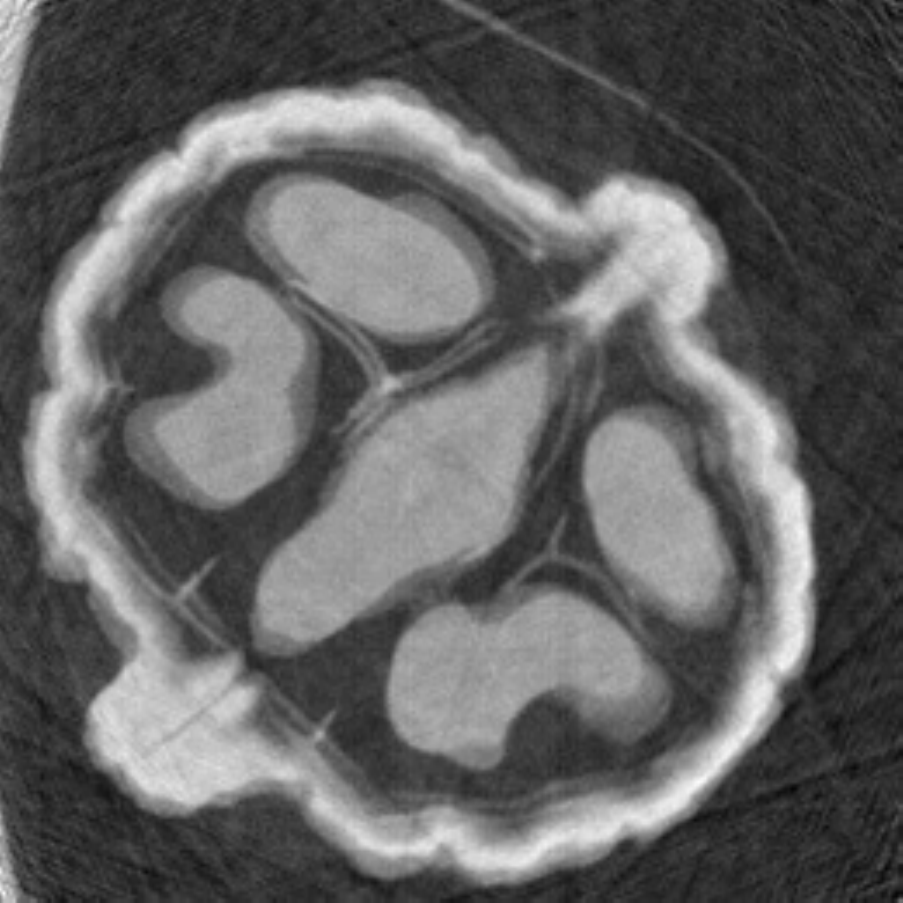}
        \caption{}
    \end{subfigure}
    \begin{subfigure}[b]{0.24\linewidth}
        \includegraphics[width=\textwidth]{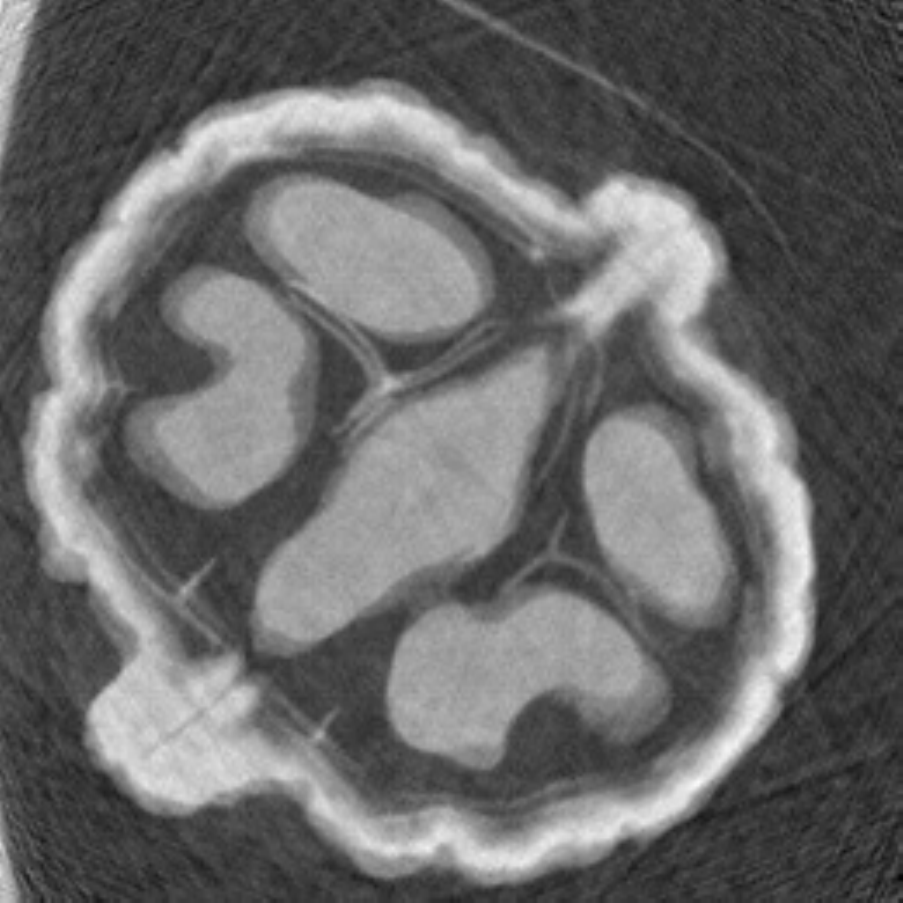}
        \caption{}
     \end{subfigure}
\par\bigskip 
    \begin{subfigure}[b]{0.24\linewidth}
        \includegraphics[width=\textwidth]{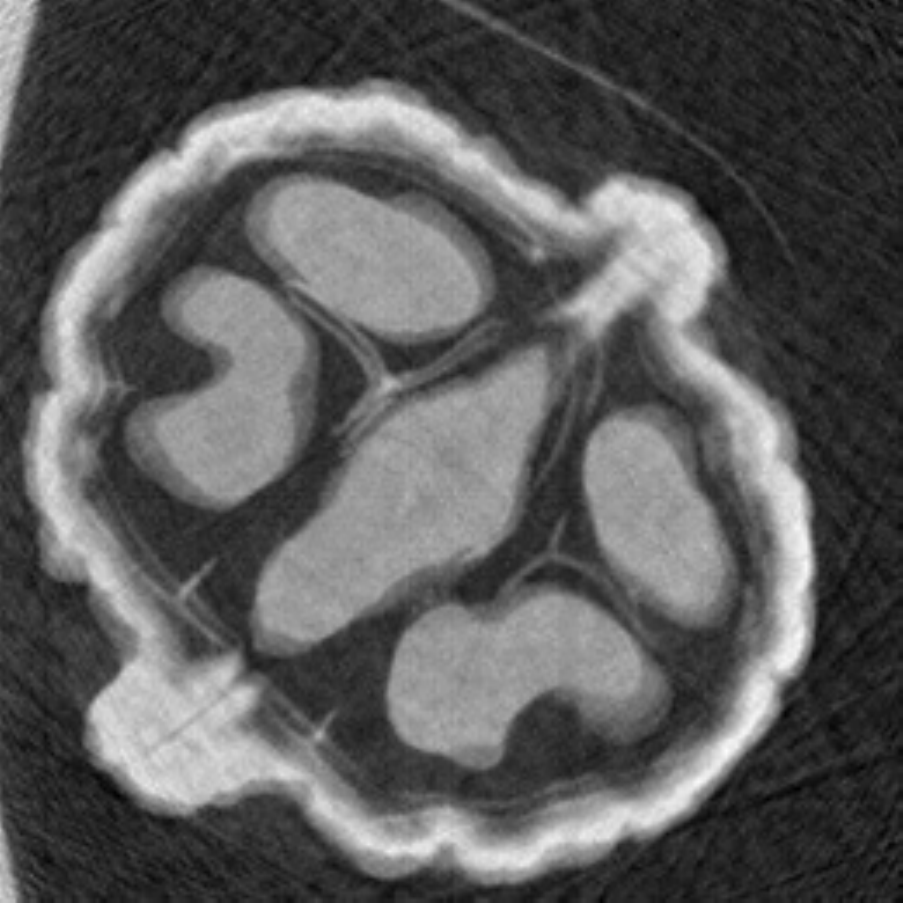}
        \caption{}
     \end{subfigure}
    \begin{subfigure}[b]{0.24\linewidth}
        \includegraphics[width=\textwidth]{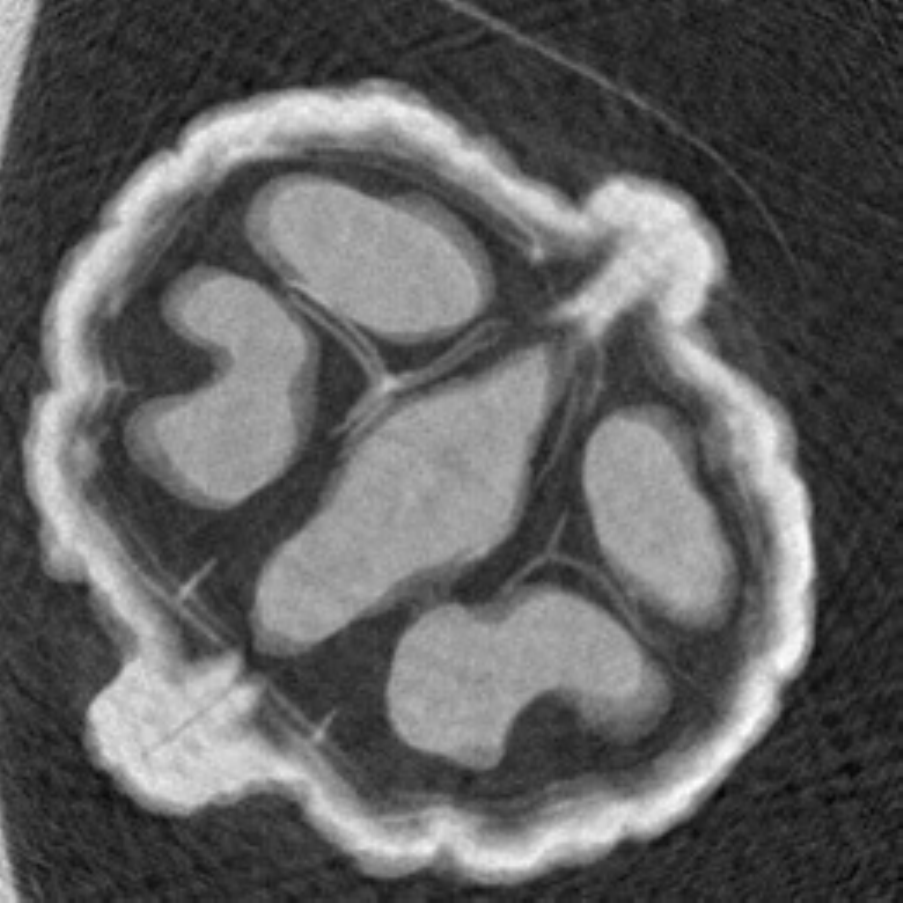}
        \caption{}
     \end{subfigure}
    \begin{subfigure}[b]{0.24\linewidth}
        \includegraphics[width=\textwidth]{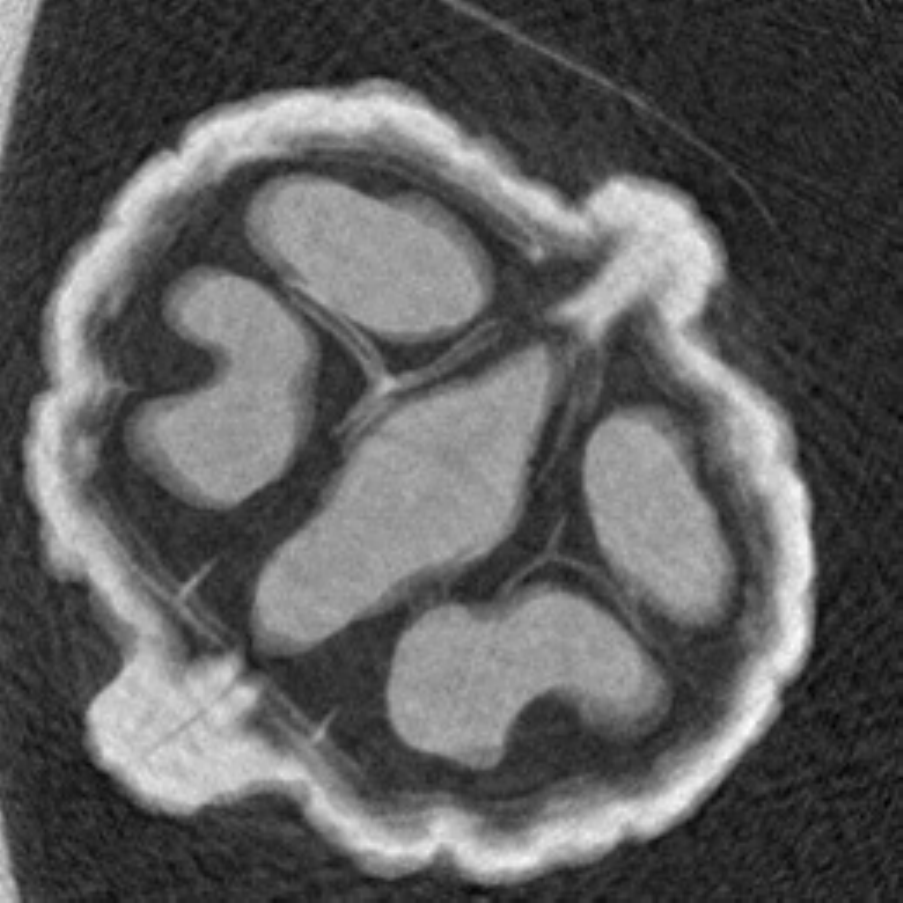}
        \caption{}
     \end{subfigure}
    \begin{subfigure}[b]{0.24\linewidth}
        \includegraphics[width=\textwidth]{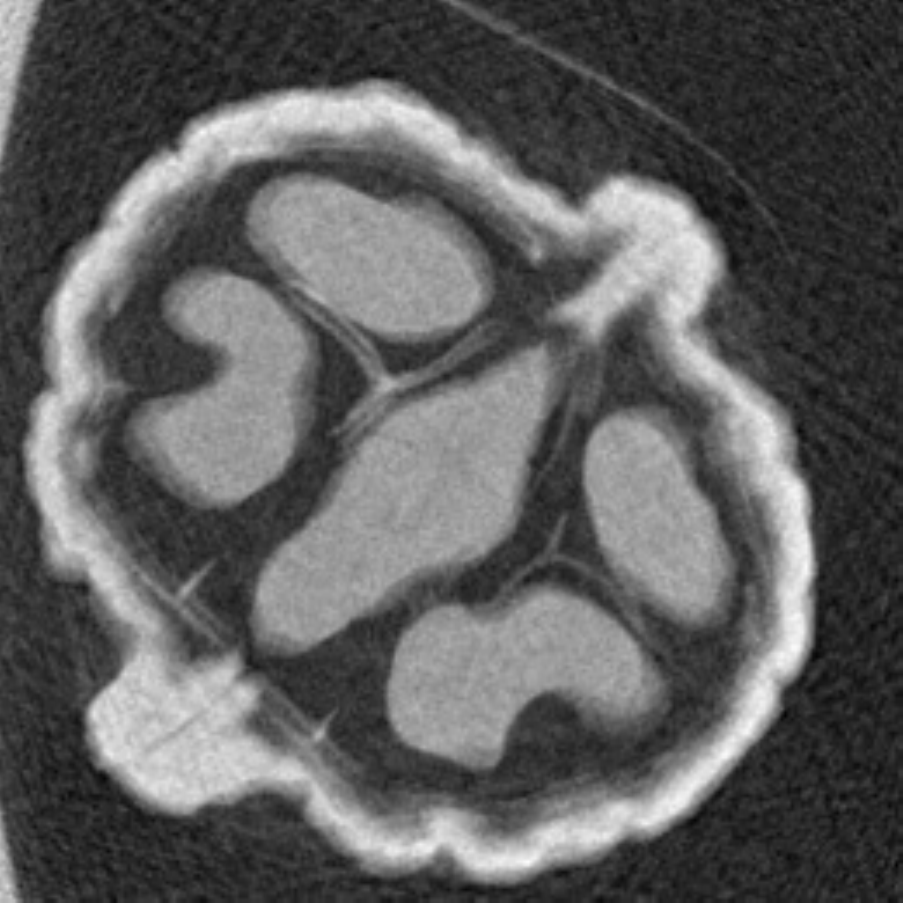}
        \caption{}
     \end{subfigure}
    \caption{Reconstructions using the global PCA based prior for various measurement angles: (a) 8 angles (b) 10 angles (c) 12 angles (d) 14 angles (e) 16 angles (f) 20 angles (g) 30 angles (h) 40 angles. $2\%$ Gaussian iid noise was added to the measurements in all the cases.}
\label{fig:all_angles}
\end{figure}

\begin{figure}
\centering
    \begin{subfigure}[b]{0.45\linewidth}
        \includegraphics[width=\textwidth]{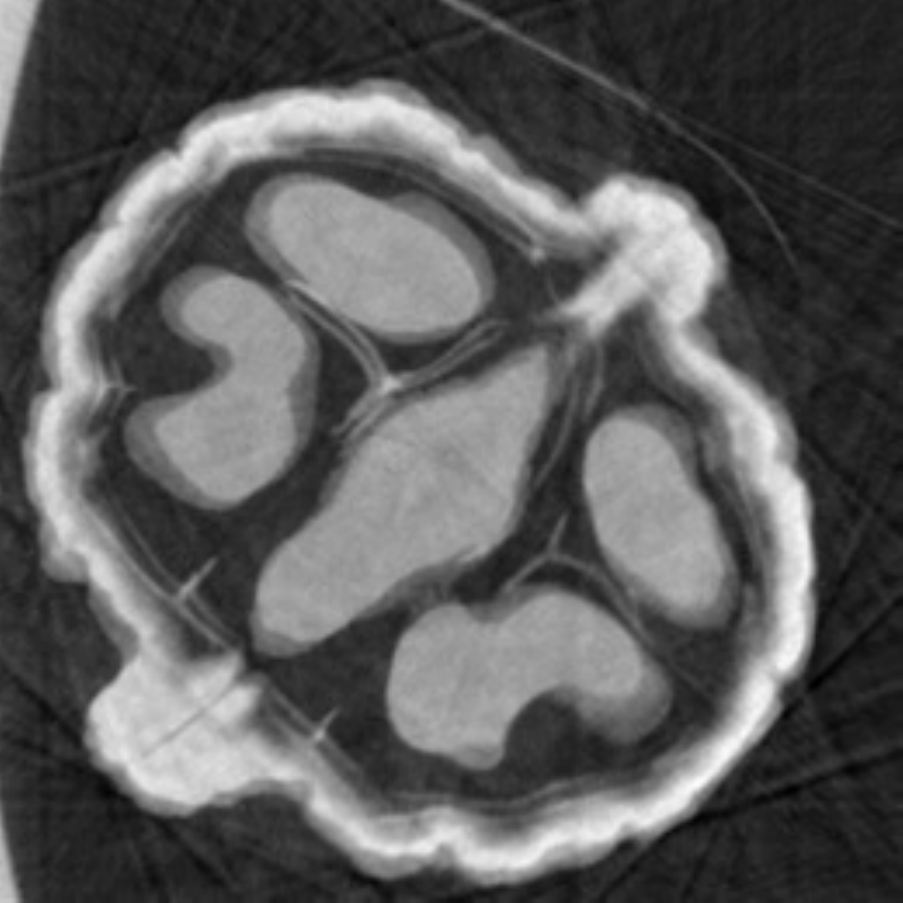}
        \caption{}
     \end{subfigure}
    \begin{subfigure}[b]{0.45\linewidth}
        \includegraphics[width=\textwidth]{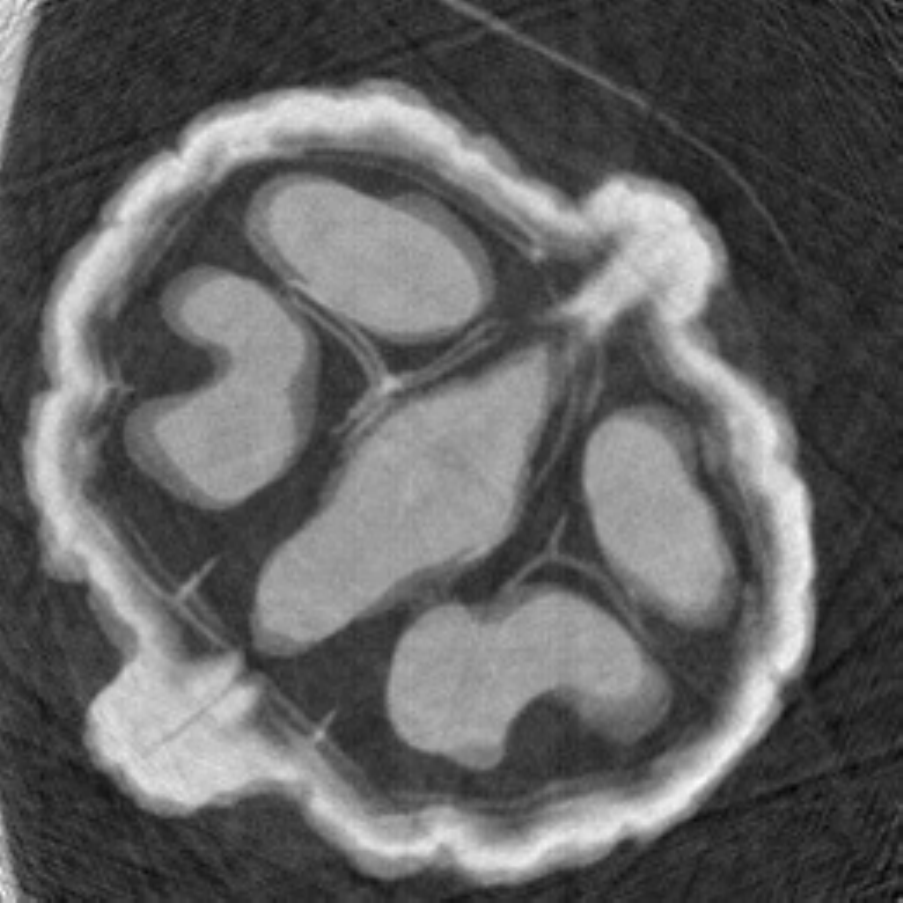}
        \caption{}
     \end{subfigure}
    \begin{subfigure}[b]{0.45\linewidth}
        \includegraphics[width=\textwidth]{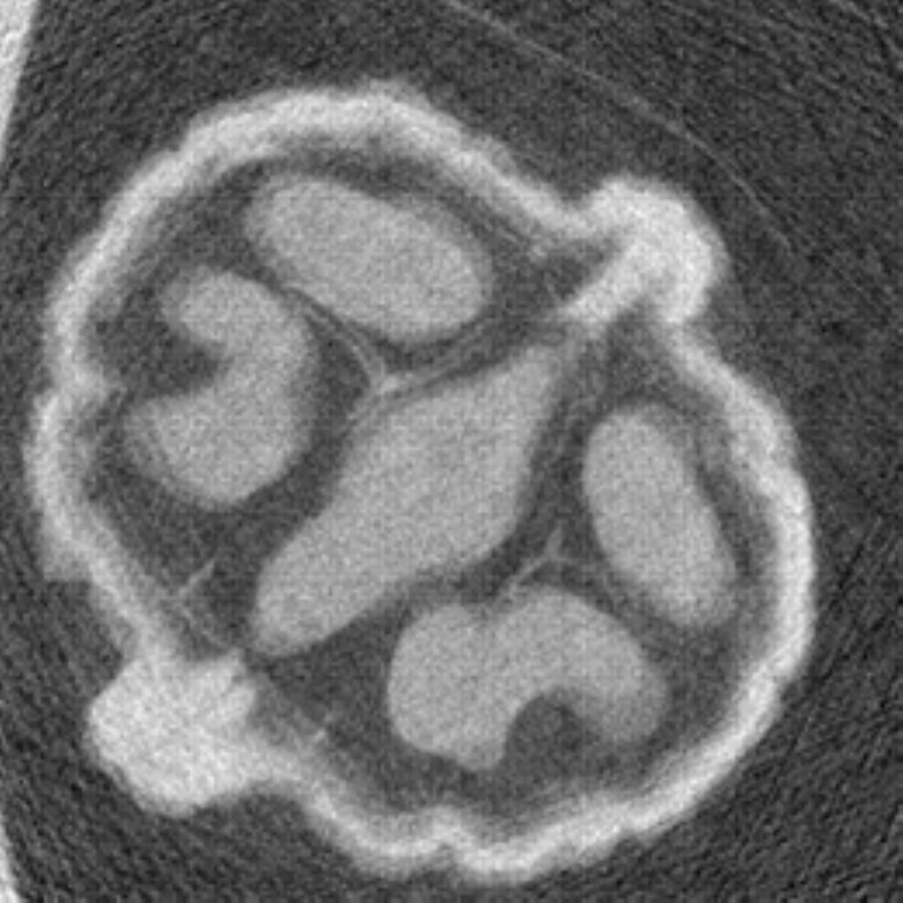}
        \caption{}
     \end{subfigure}
    \caption{Performance of the global PCA based reconstruction for
      different noise levels (measurements for 12 angles): (a) no
      noise (relative mean squared error (relative mse) = 0.0331, ssim =
      0.3475), (b) $2\%$ noise
      (relative mse = 0.0370, ssim = 0.3097) (h) $10\%$ noise (relative mse =
      0.0697, ssim = 0.2191).} 
\label{fig:diff_noise_levels}
\end{figure}

 Figures ~\ref{fig:all_angles} and~\ref{fig:diff_noise_levels} show the results of PCA based reconstruction for different measurement angles and different levels of noise, after tuning for the optimal $\lambda_1$ and $\lambda_2$.

\begin{table}[]
  \centering
\setlength\extrarowheight{5.5pt}
\caption{Performance of the global PCA based reconstruction for various measurement angles, corresponding to the results shown in figure~\ref{fig:all_angles}}
\label{my-label}
\begin{tabular}{|l|l|l|}
  \hline
\textbf{\begin{tabular}[c]{@{}l@{}}No. of angles\\ (uniformly distributed)\end{tabular}} & \textbf{relative mse} & \textbf{ssim} \\ \hline
8 & 0.0389 & 0.3113 \\ \hline
10 & 0.0379 & 0.3165 \\ \hline
12 & 0.0370 & 0.3097 \\ \hline
14 & 0.0366 & 0.3111 \\ \hline
16 & 0.0347 & 0.3163 \\ \hline
20 & 0.0330 & 0.3248 \\ \hline
30 & 0.0300 & 0.3351 \\ \hline
40 & 0.0275 & 0.3472 \\ \hline
\end{tabular}
\end{table}

\begin{figure}
\centering
    \begin{subfigure}[b]{0.49\linewidth}
        \includegraphics[width=\textwidth]{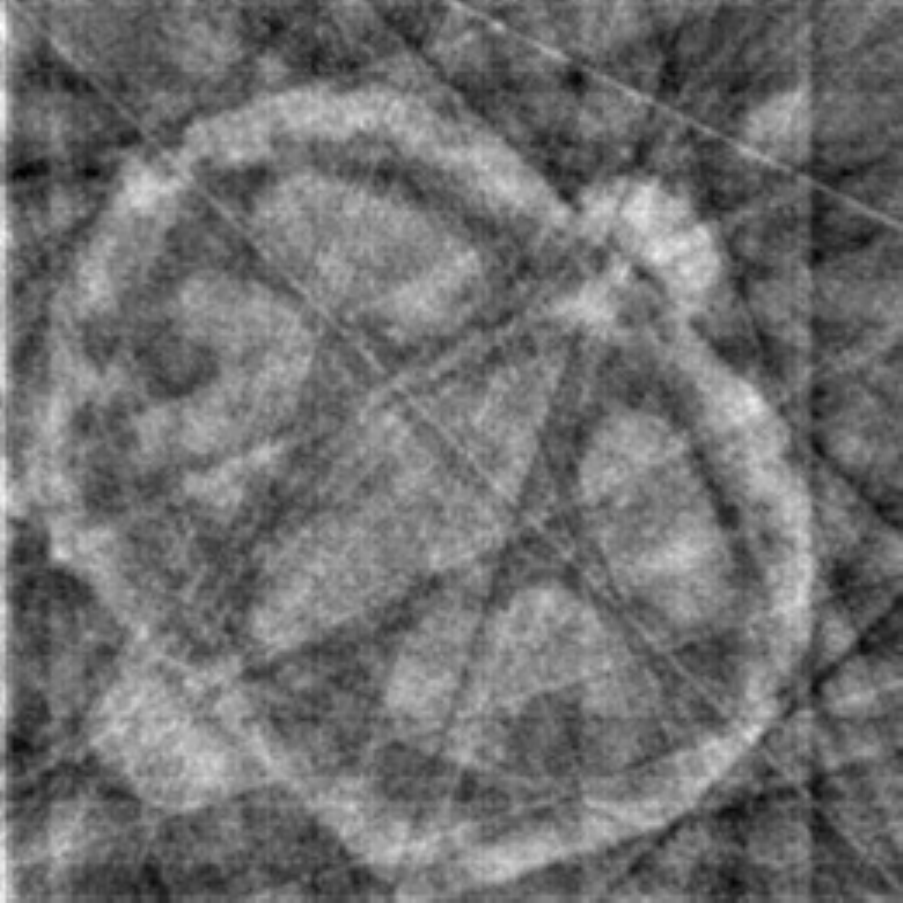}
        \caption{}
     \end{subfigure}
    \begin{subfigure}[b]{0.49\linewidth}
        \includegraphics[width=\textwidth]{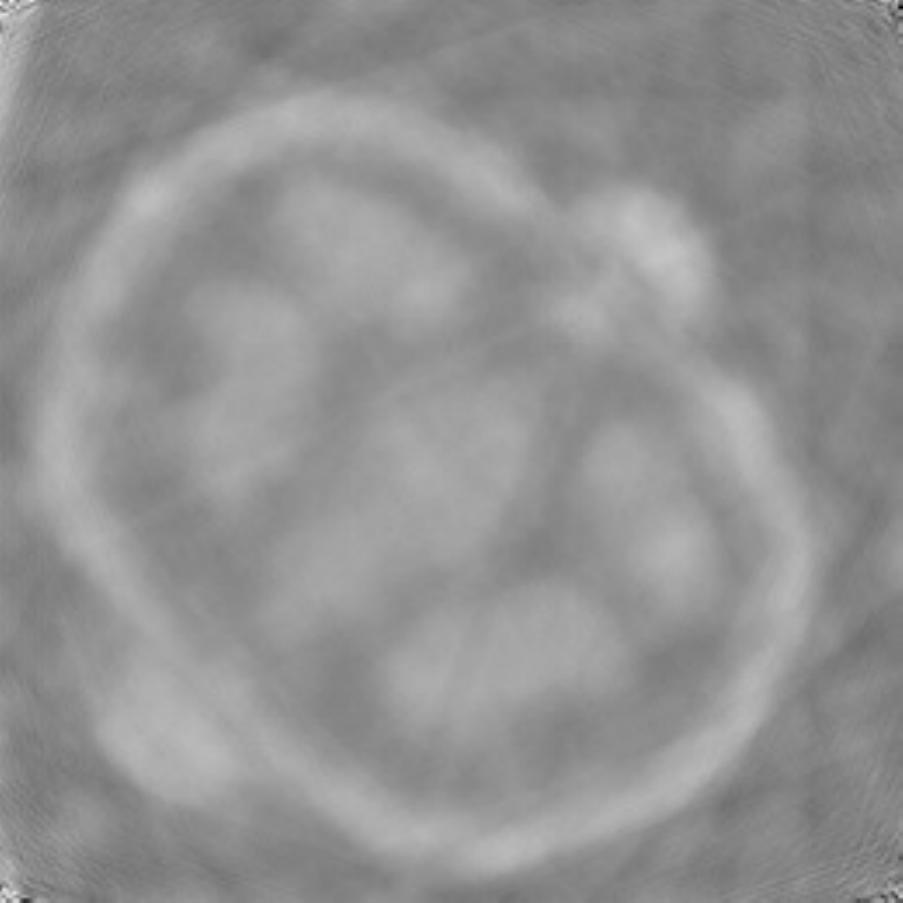}
        \caption{}
    \end{subfigure}
\par\bigskip 
    \begin{subfigure}[b]{0.49\linewidth}
        \includegraphics[width=\textwidth]{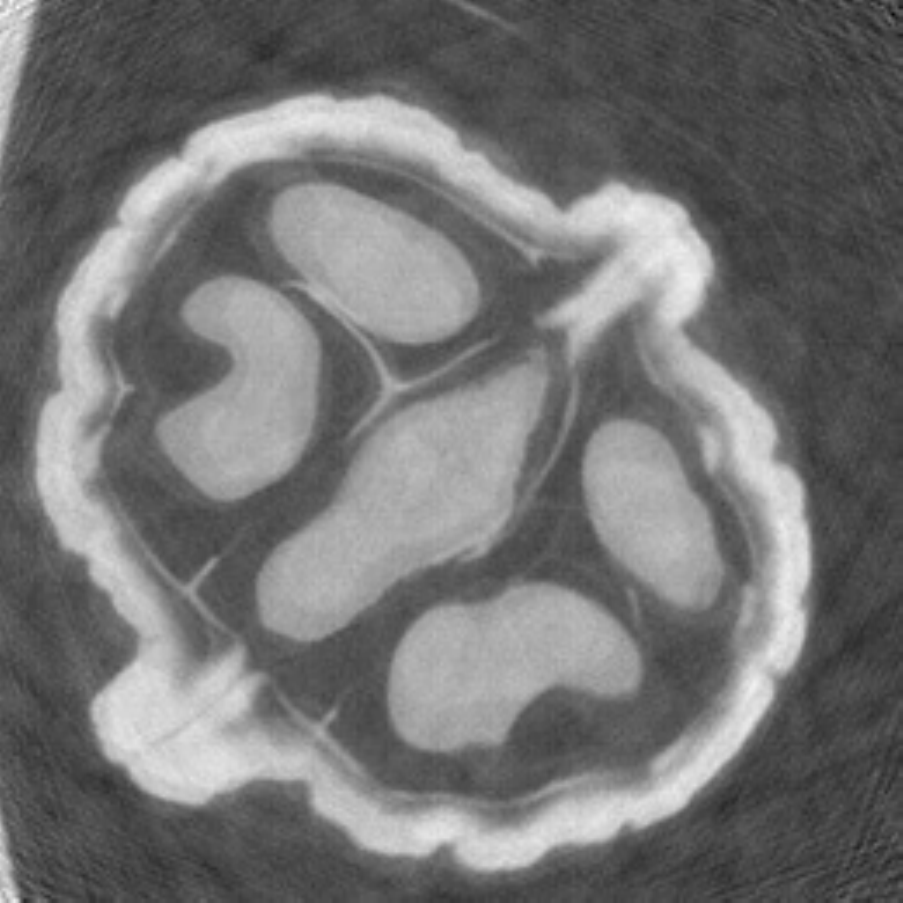}
        \caption{}
     \end{subfigure}
    \begin{subfigure}[b]{0.49\linewidth}
        \includegraphics[width=\textwidth]{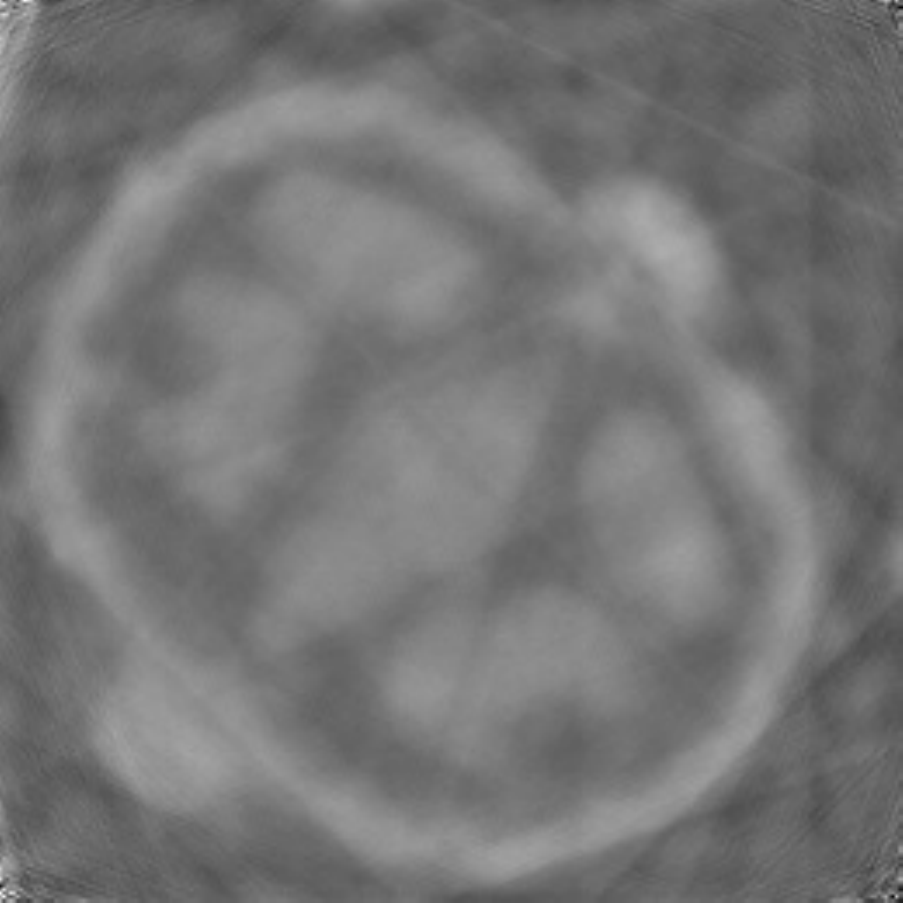}
        \caption{}
     \end{subfigure}
    \caption{Reconstruction of a slice from the walnut CT dataset~\cite{walnut} from measurements along 12 angles and with 2\% noise. (a) Reconstructed by filtered backprojection (b) Reconstructed slice using plain CS, without any prior (c) Reconstructed slice using CS, with global prior (d) Reconstructed slice using CS, with patch based prior with patch size = $13\times 13$ pixels.}
\label{fig:global_local_walnut}
\end{figure}

\begin{figure}
\centering
\begin{subfigure}[b]{0.99\linewidth}
        \includegraphics[width=\linewidth]{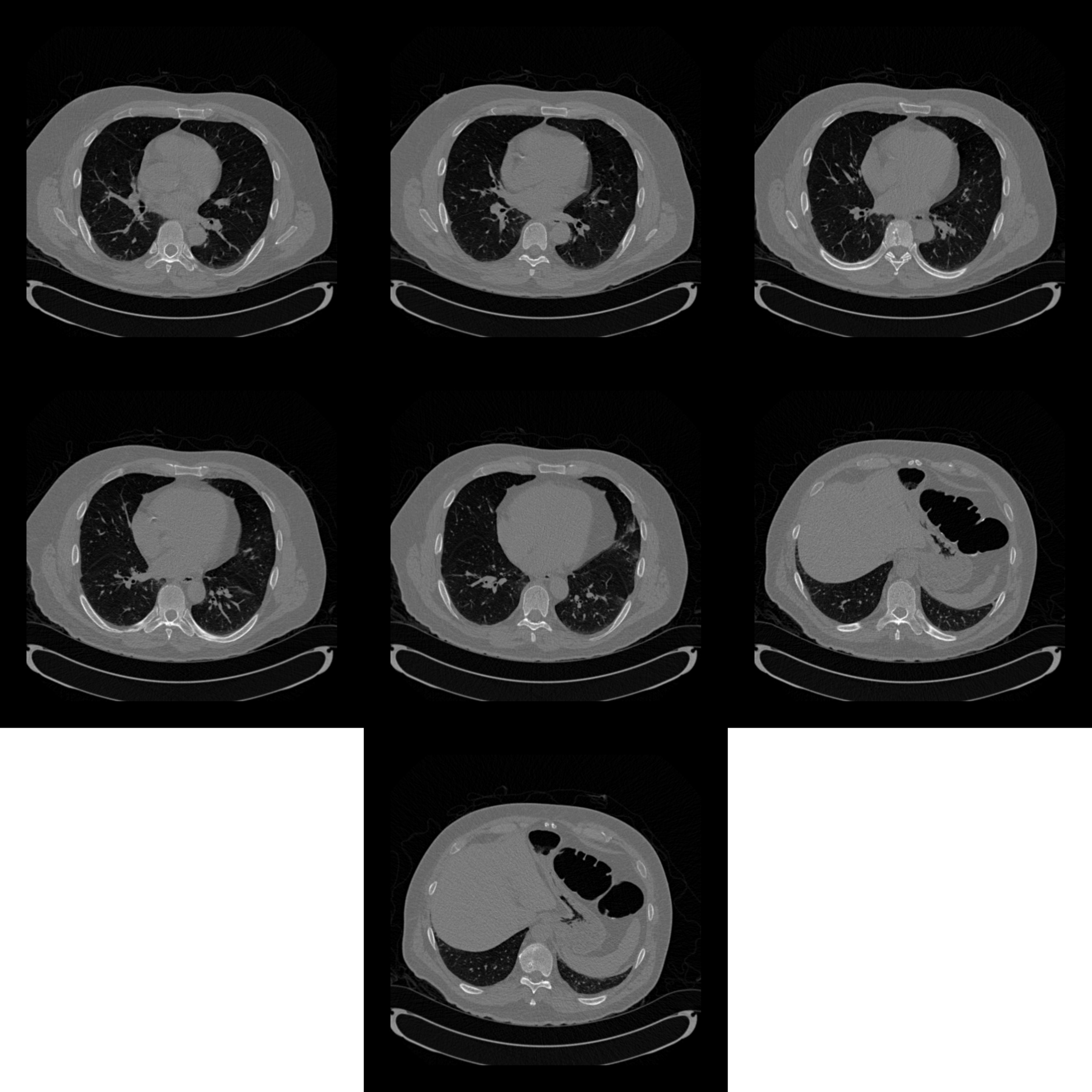}
\caption{}
\label{fig:colon_templates}
\end{subfigure}
    \begin{subfigure}[b]{0.33\linewidth}
        \includegraphics[width=\textwidth]{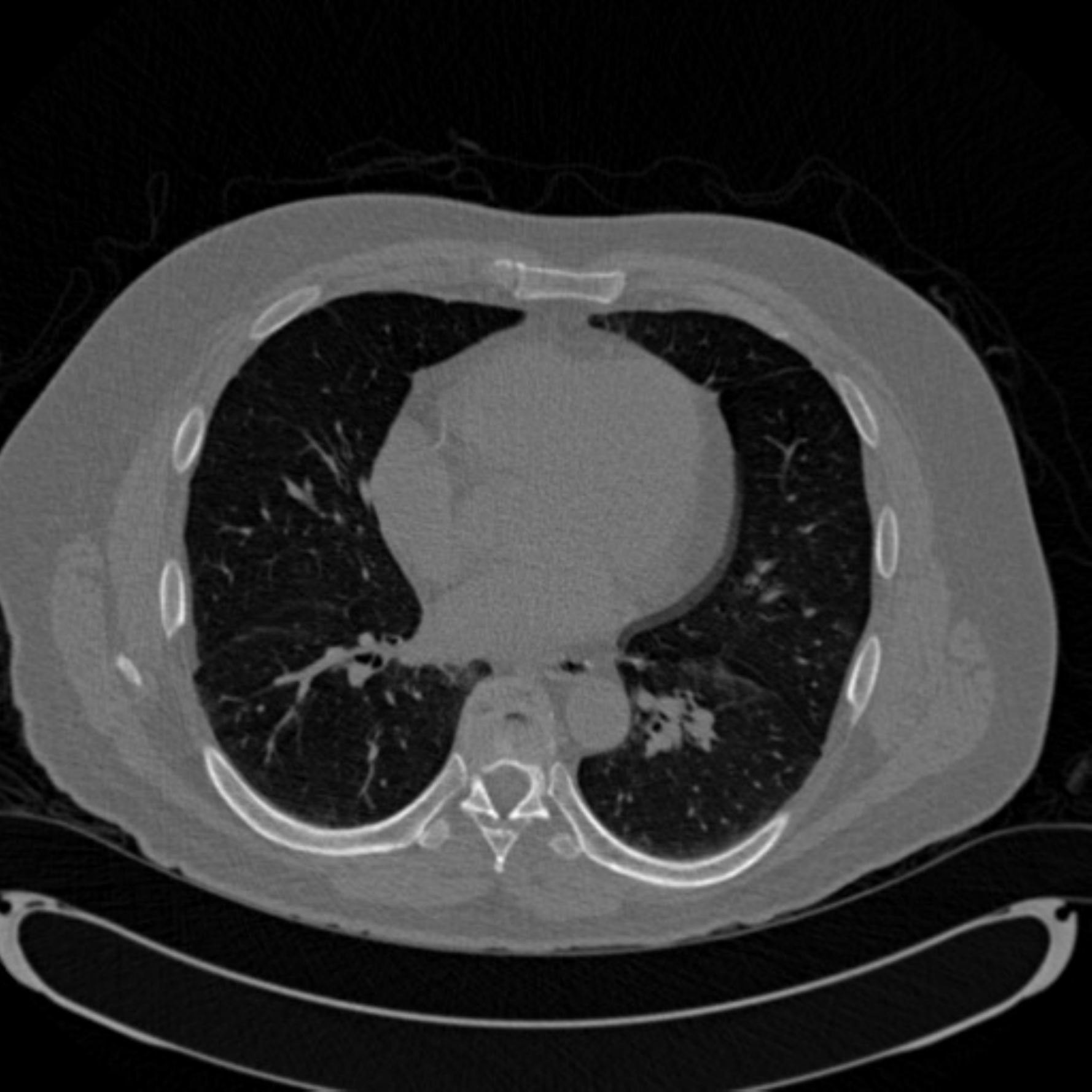}
        \caption{}
\label{fig:colon_test}
     \end{subfigure}
        \caption{Data from colon CT dataset~\cite{colon}. Each slice is of size $512\times 512$. (a) Seven templates (slice nos. 13, 23, 33, 43, 53, 63 and 73 from the CT volume) that were used to build the prior. (b) The test data (slice no. 38 from the same volume).}

\end{figure}

\begin{figure}
\centering
    \begin{subfigure}[b]{0.45\linewidth}
        \includegraphics[width=\textwidth]{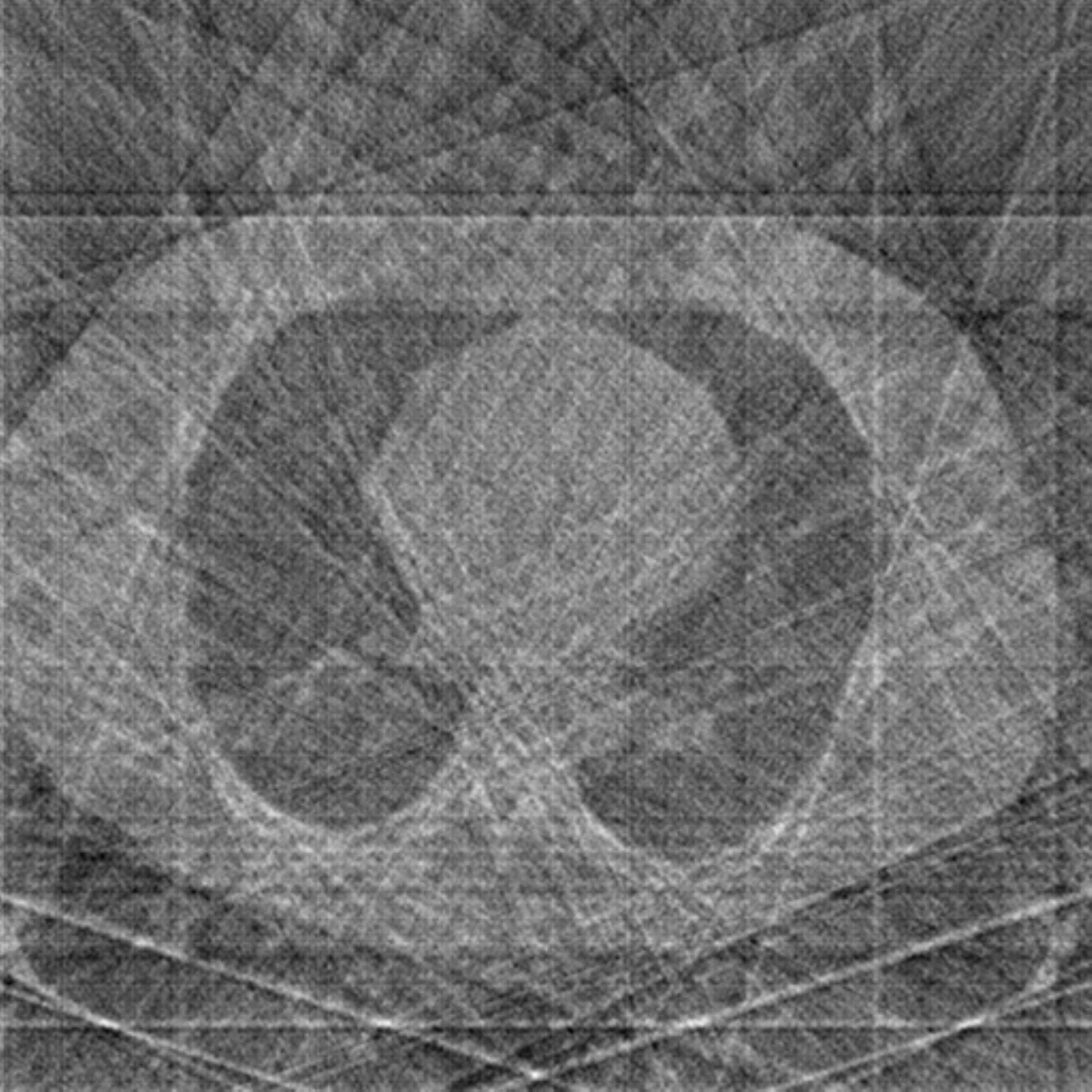}
        \caption{}
     \end{subfigure}
    \begin{subfigure}[b]{0.45\linewidth}
        \includegraphics[width=\textwidth]{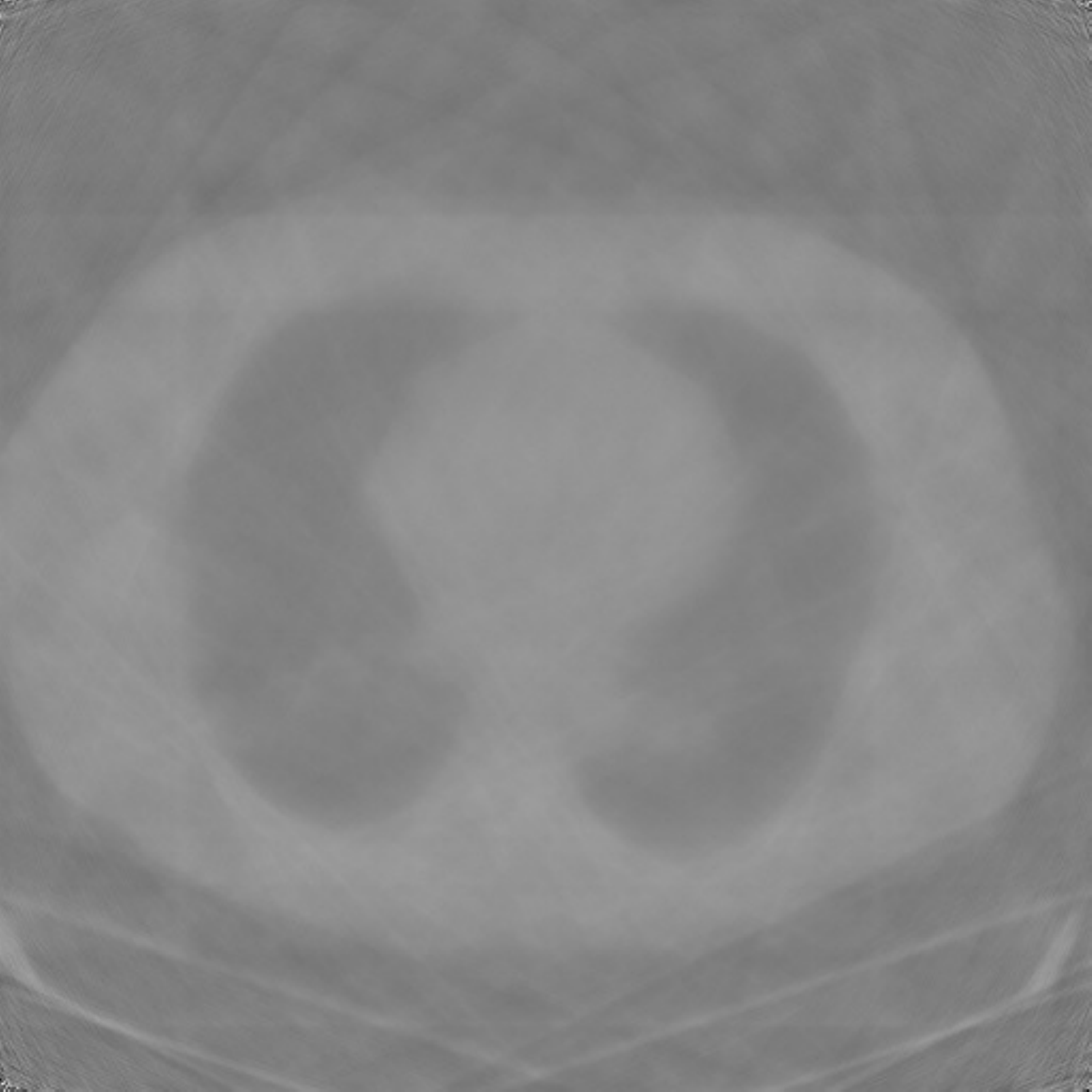}
        \caption{}

    \end{subfigure}
\par\bigskip 
    \begin{subfigure}[b]{0.45\linewidth}
        \includegraphics[width=\textwidth]{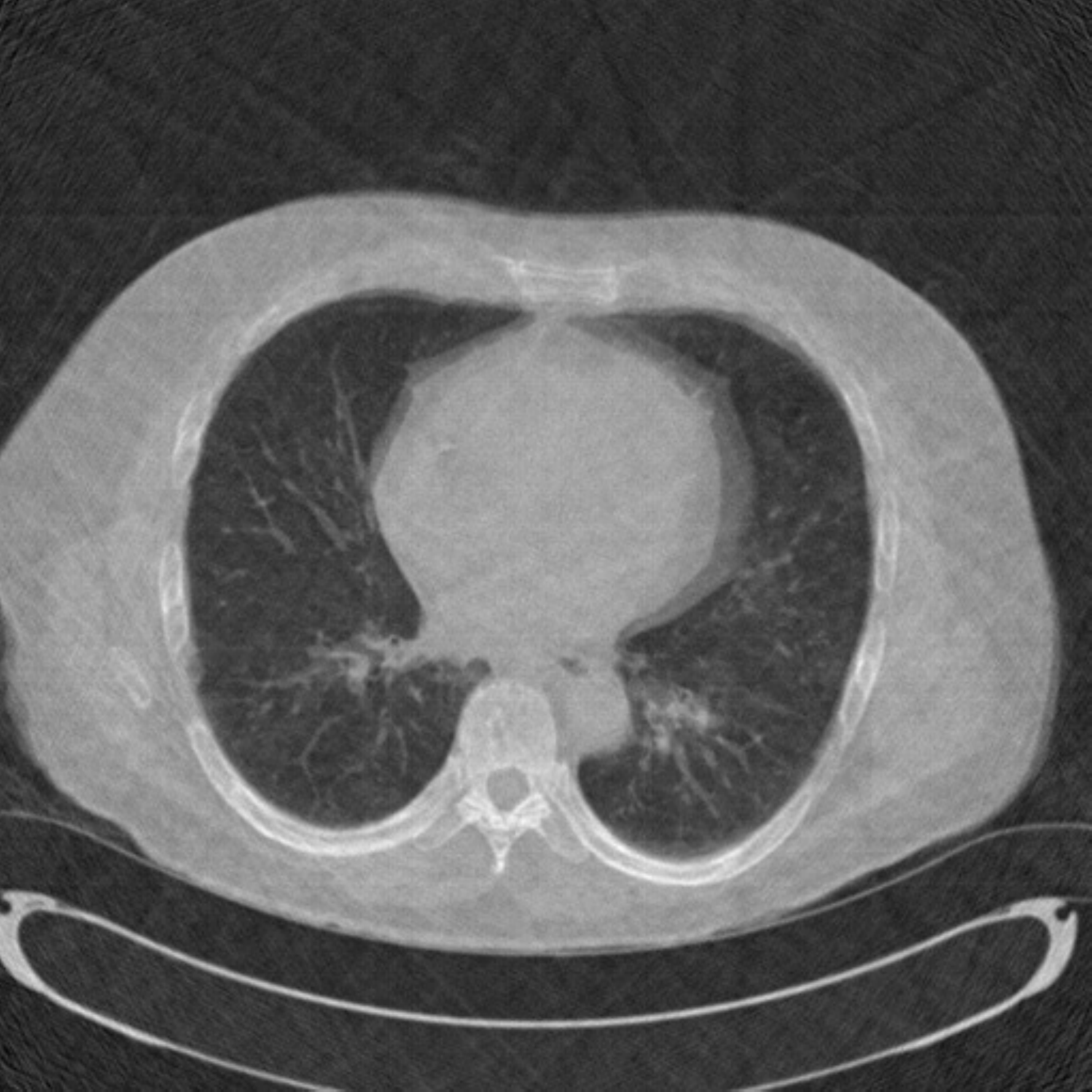}
        \caption{}
     \end{subfigure}
    \begin{subfigure}[b]{0.45\linewidth}
        \includegraphics[width=\textwidth]{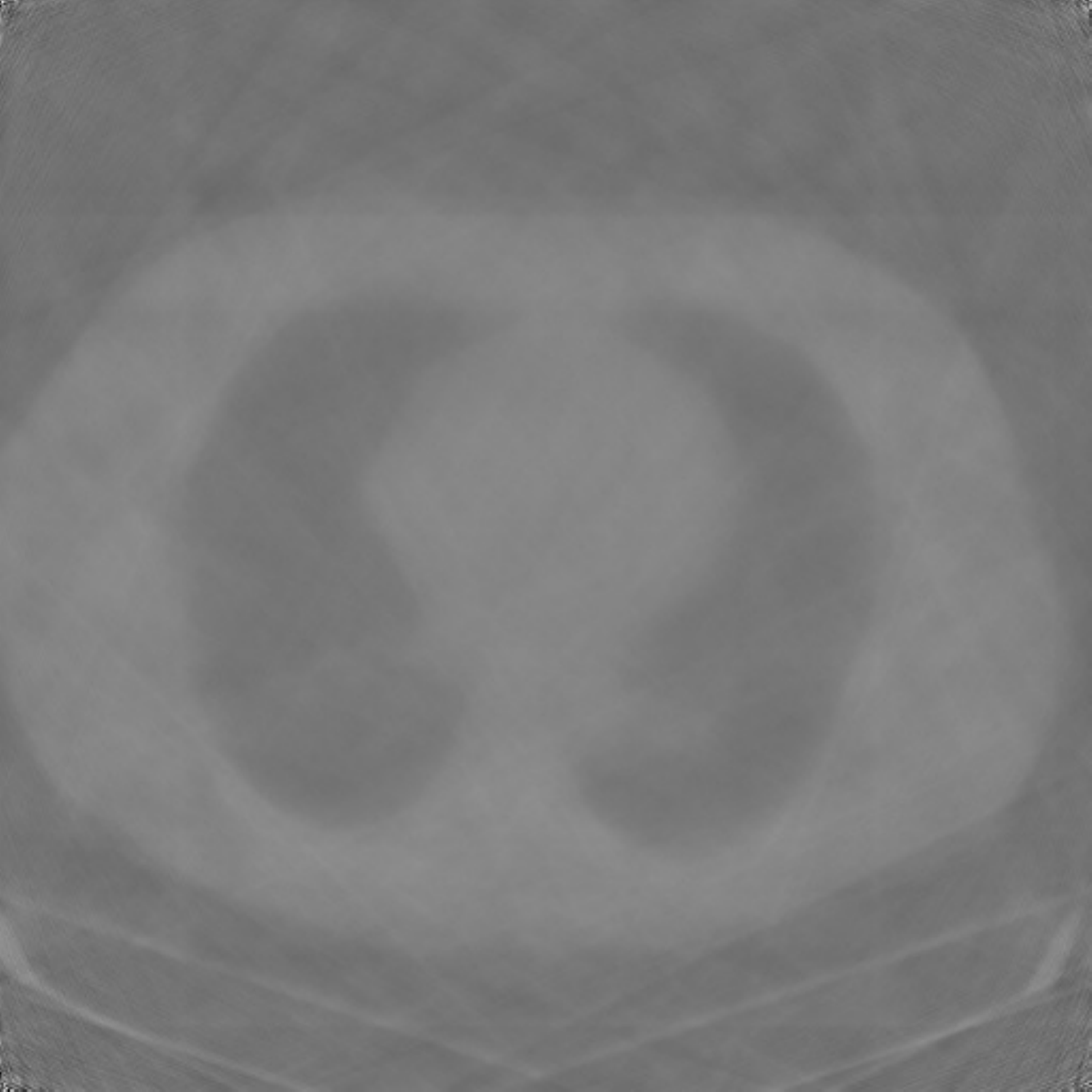}
        \caption{}
     \end{subfigure}
    \caption{Reconstruction of a slice from the colon CT dataset~\cite{colon} from measurements along 15 angles and with 2\% noise. (a) Reconstructed by filtered backprojection (b) Reconstructed slice using plain CS, without any prior (c) Reconstructed slice using CS, with global prior (d) Reconstructed slice using CS, with patch based prior with patch size = $8\times 8$ pixels.}
\label{fig:global_local_colon}
\end{figure}

\begin{figure}
\centering
\begin{subfigure}[b]{\linewidth}
        \includegraphics[width=\linewidth]{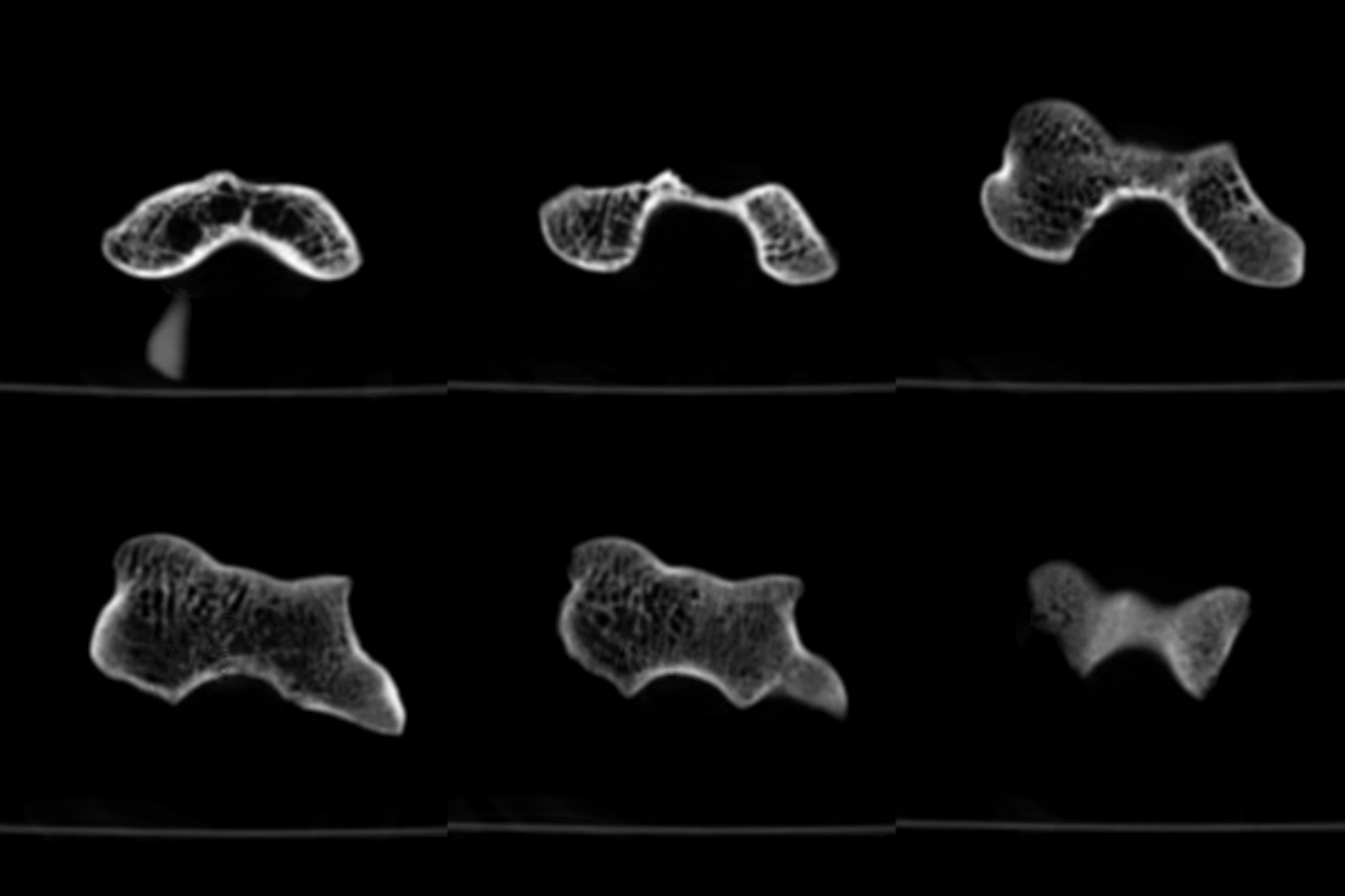}
\caption{}
\label{fig:humerus_templates}
\end{subfigure}
    \begin{subfigure}[b]{0.33\linewidth}
        \includegraphics[width=\textwidth]{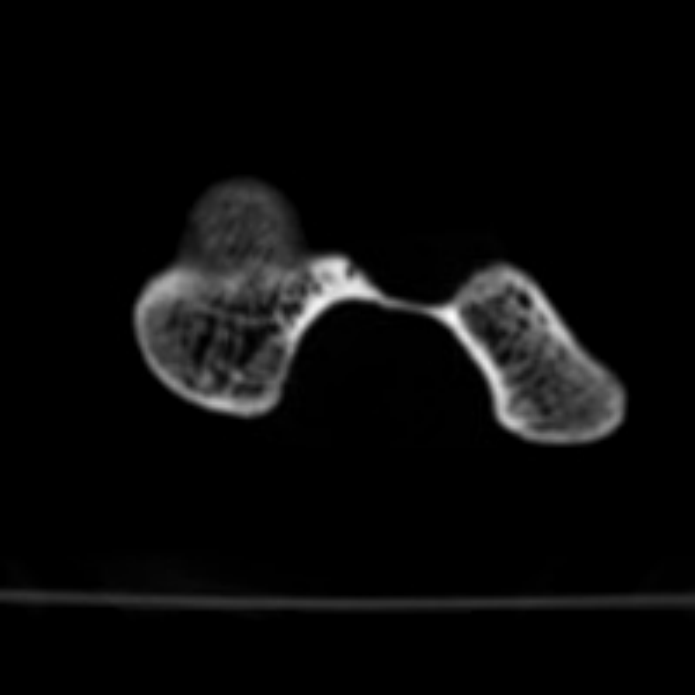}
        \caption{}
\label{fig:humerus_test}
     \end{subfigure}
        \caption{Data from humerus CT dataset~\cite{humerus}. Each slice is of size $200\times 200$. (a) Six templates (slice nos. 55, 65, 75, 85, 95 and 105 from the CT volume) that were used to build the prior. (b) The test data (slice no. 70 from the same volume).}

\end{figure}

\begin{figure}
\centering
    \begin{subfigure}[b]{0.45\linewidth}
        \includegraphics[width=\textwidth]{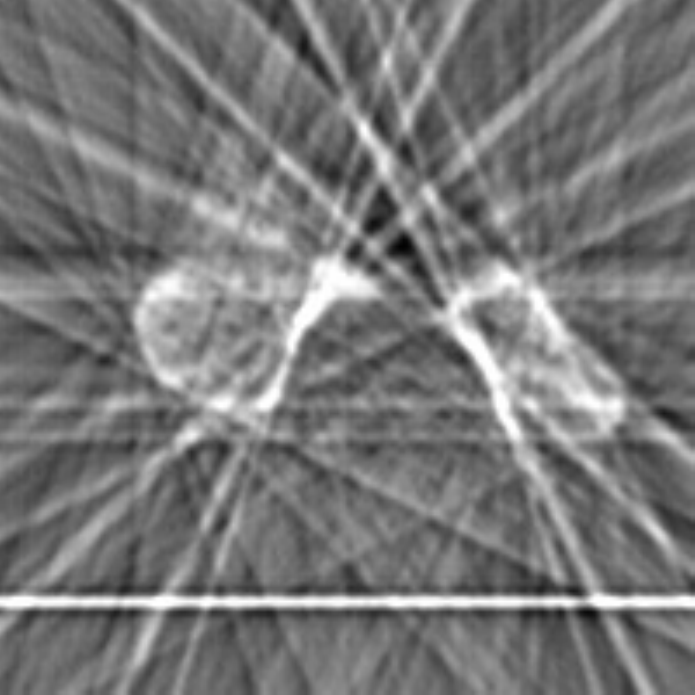}
        \caption{}
     \end{subfigure}
    \begin{subfigure}[b]{0.45\linewidth}
        \includegraphics[width=\textwidth]{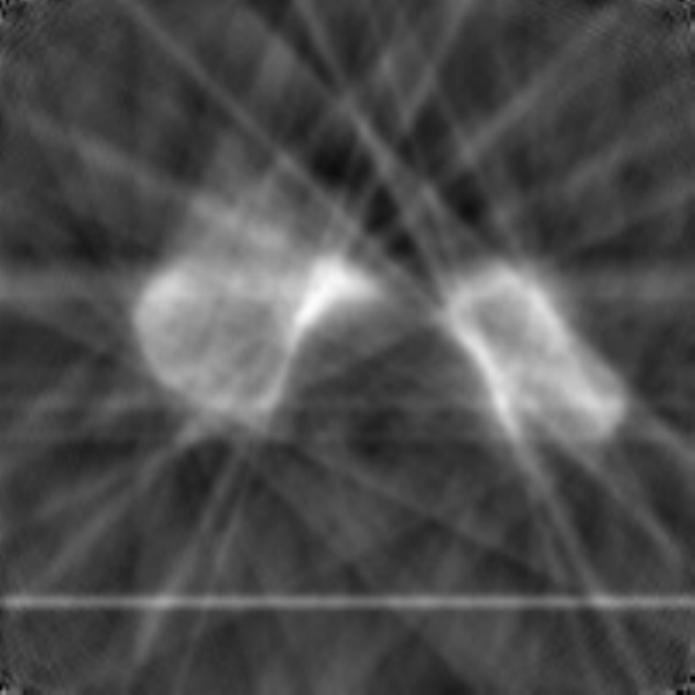}
        \caption{}
    \end{subfigure}
\par\bigskip 
    \begin{subfigure}[b]{0.45\linewidth}
        \includegraphics[width=\textwidth]{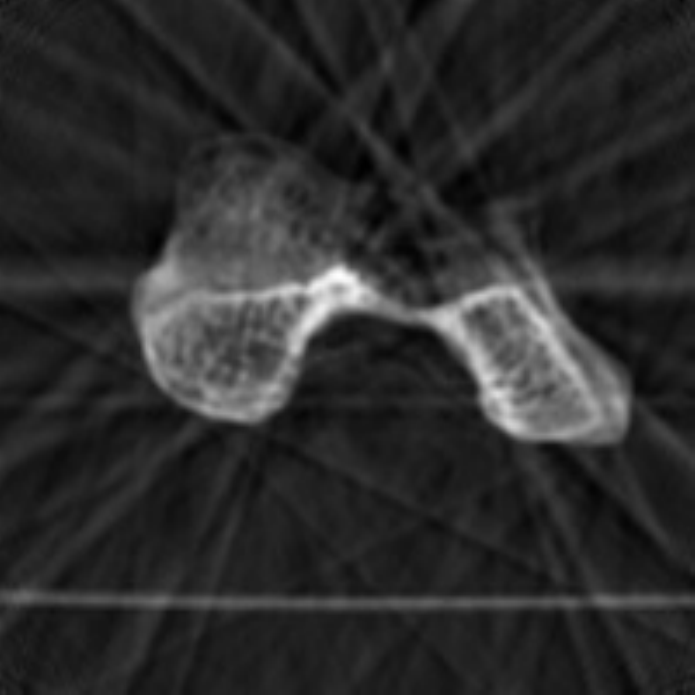}
        \caption{}
     \end{subfigure}
    \begin{subfigure}[b]{0.45\linewidth}
        \includegraphics[width=\textwidth]{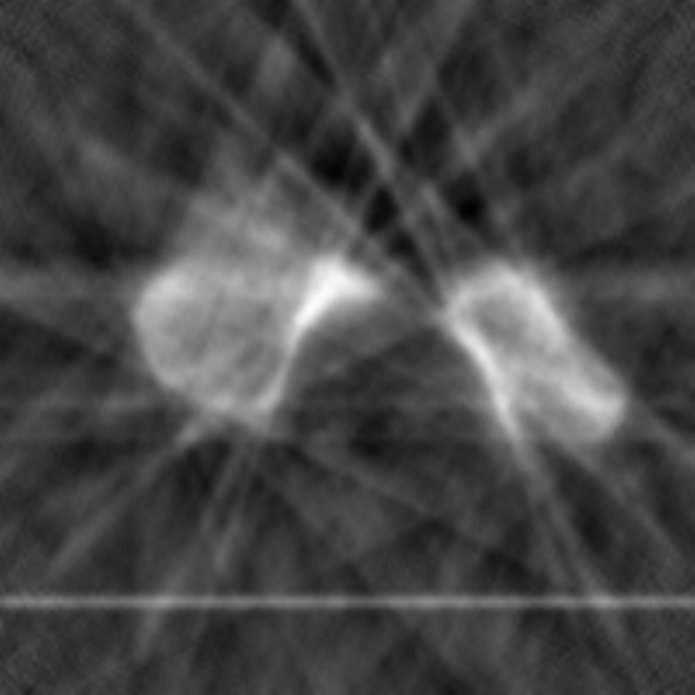}
        \caption{}
     \end{subfigure}
    \caption{Reconstruction of a slice from the humerus bone CT dataset~\cite{humerus} from measurements along 10 angles and with 2\% noise. (a) Reconstructed by filtered backprojection (b) Reconstructed slice using plain CS, without any prior (c) Reconstructed slice using CS, with global prior (d) Reconstructed slice using CS, with patch based prior with patch size = $8\times 8$ pixels.}
\label{fig:global_local_humerus}
\end{figure}

\begin{figure}
\centering
\begin{subfigure}[b]{0.85\linewidth}
        \includegraphics[width=\linewidth]{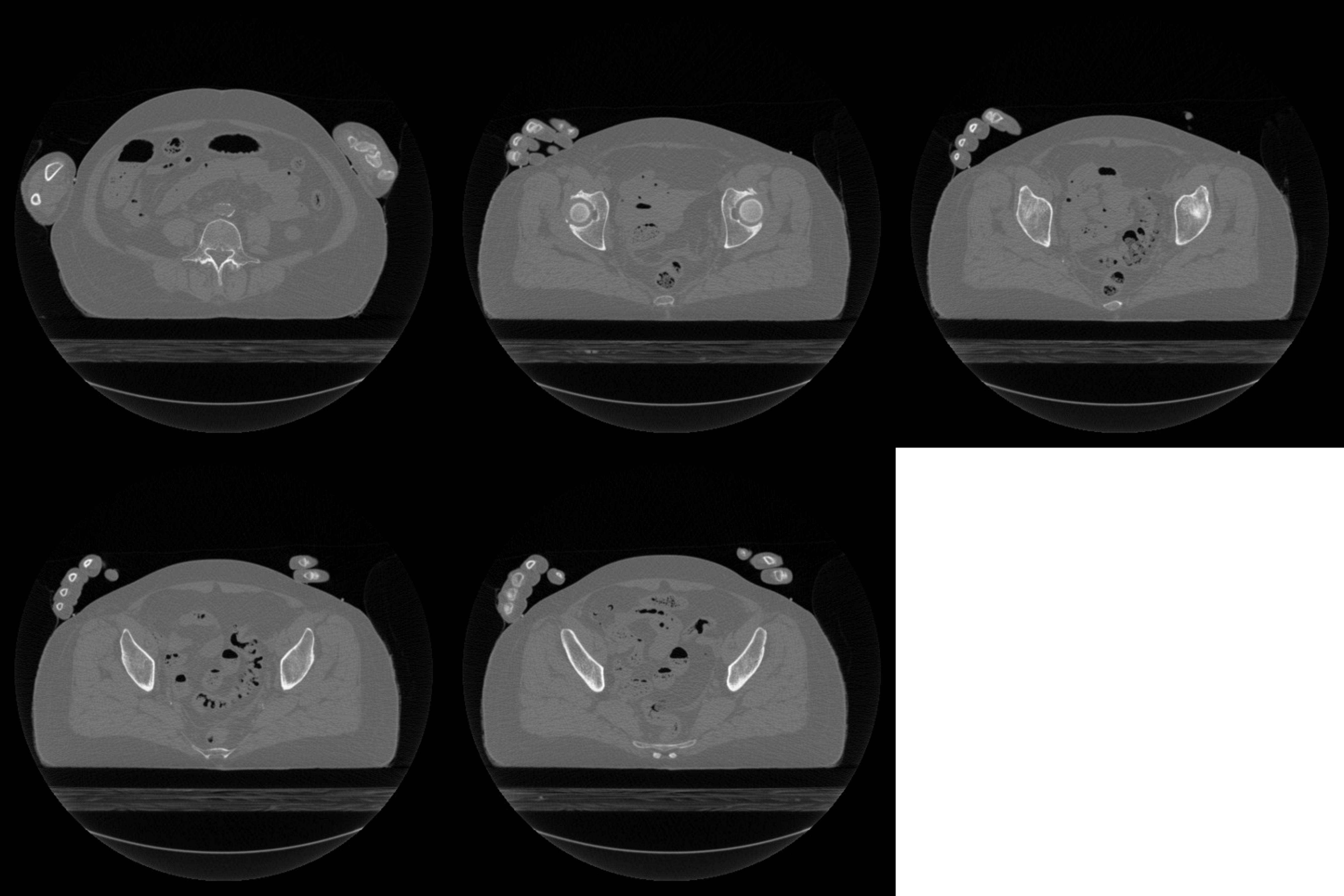}
\caption{}
\label{fig:pelvis_templates}
\end{subfigure}
    \begin{subfigure}[b]{0.3\linewidth}
        \includegraphics[width=\textwidth]{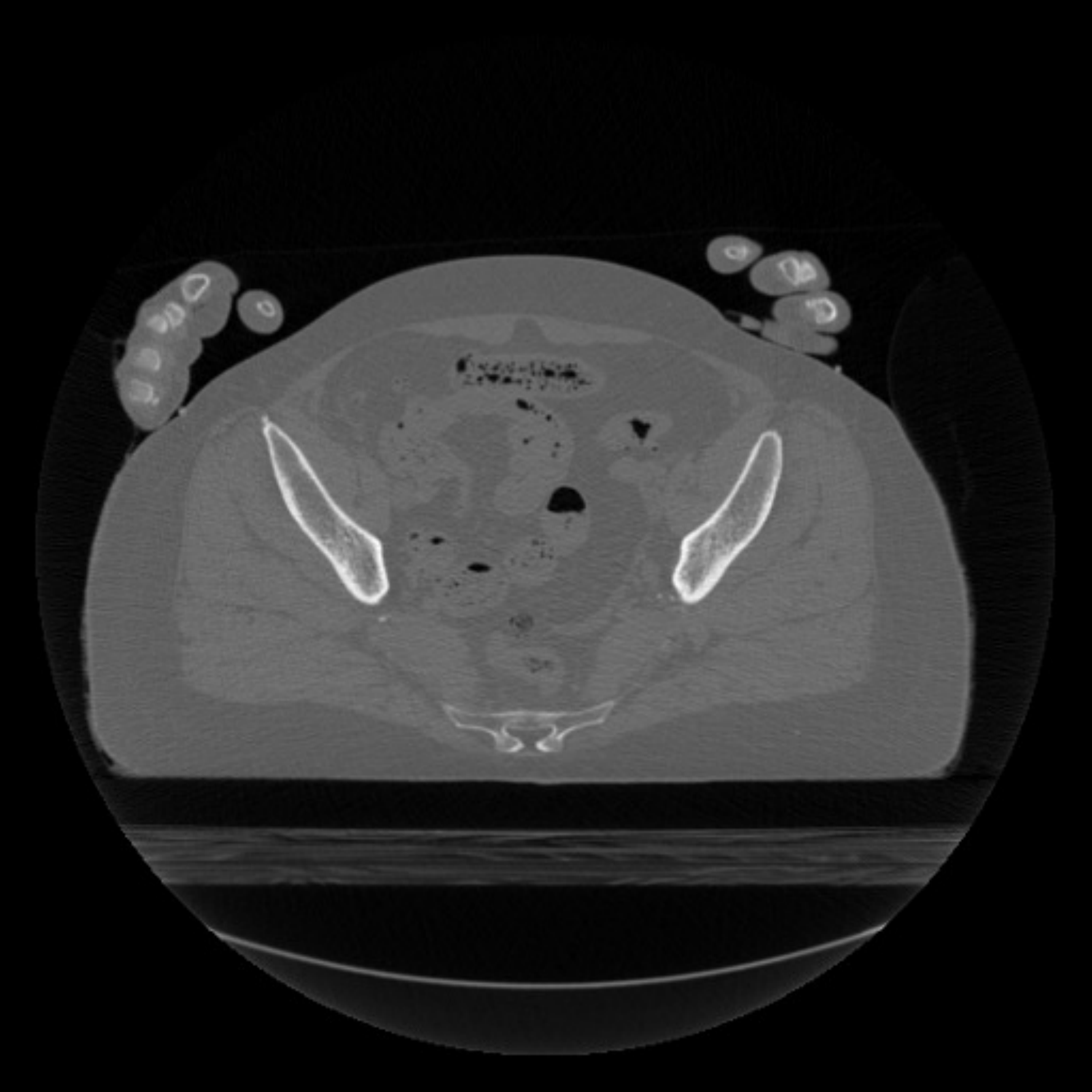}
        \caption{}
\label{fig:pelvis_test}
     \end{subfigure}
        \caption{Data from pelvis CT dataset~\cite{pelvis}. Each slice is of size $512\times 512$. (a) Five templates (slice nos. 50, 60, 70, 80 and 90 from the CT volume) that were used to build the prior. (b) The test data (slice no. 95 from the same volume).}
\end{figure}

\begin{figure}[t]
\centering
    \begin{subfigure}[b]{0.45\linewidth}
        \includegraphics[width=\textwidth]{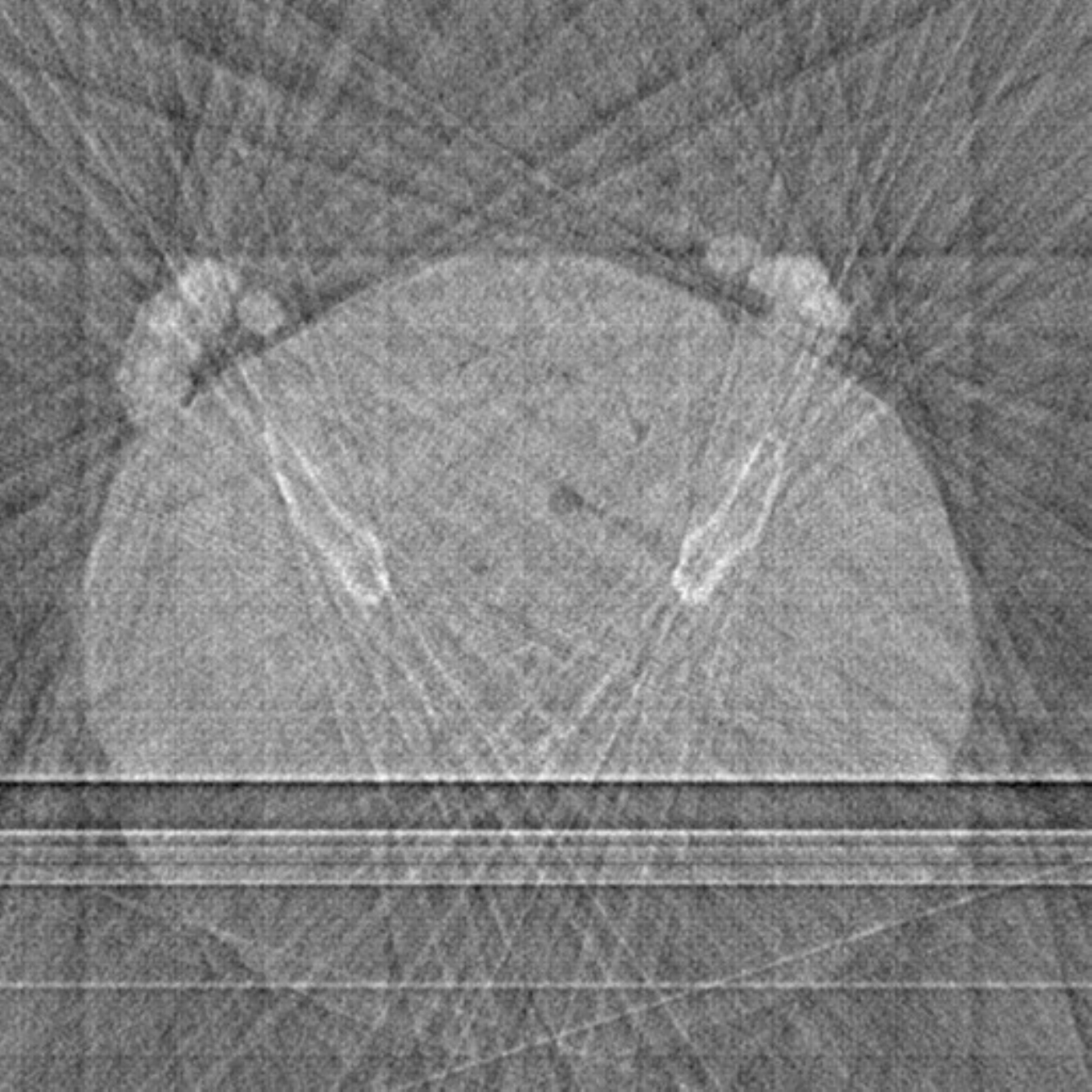}
        \caption{}
     \end{subfigure}
    \begin{subfigure}[b]{0.45\linewidth}
        \includegraphics[width=\textwidth]{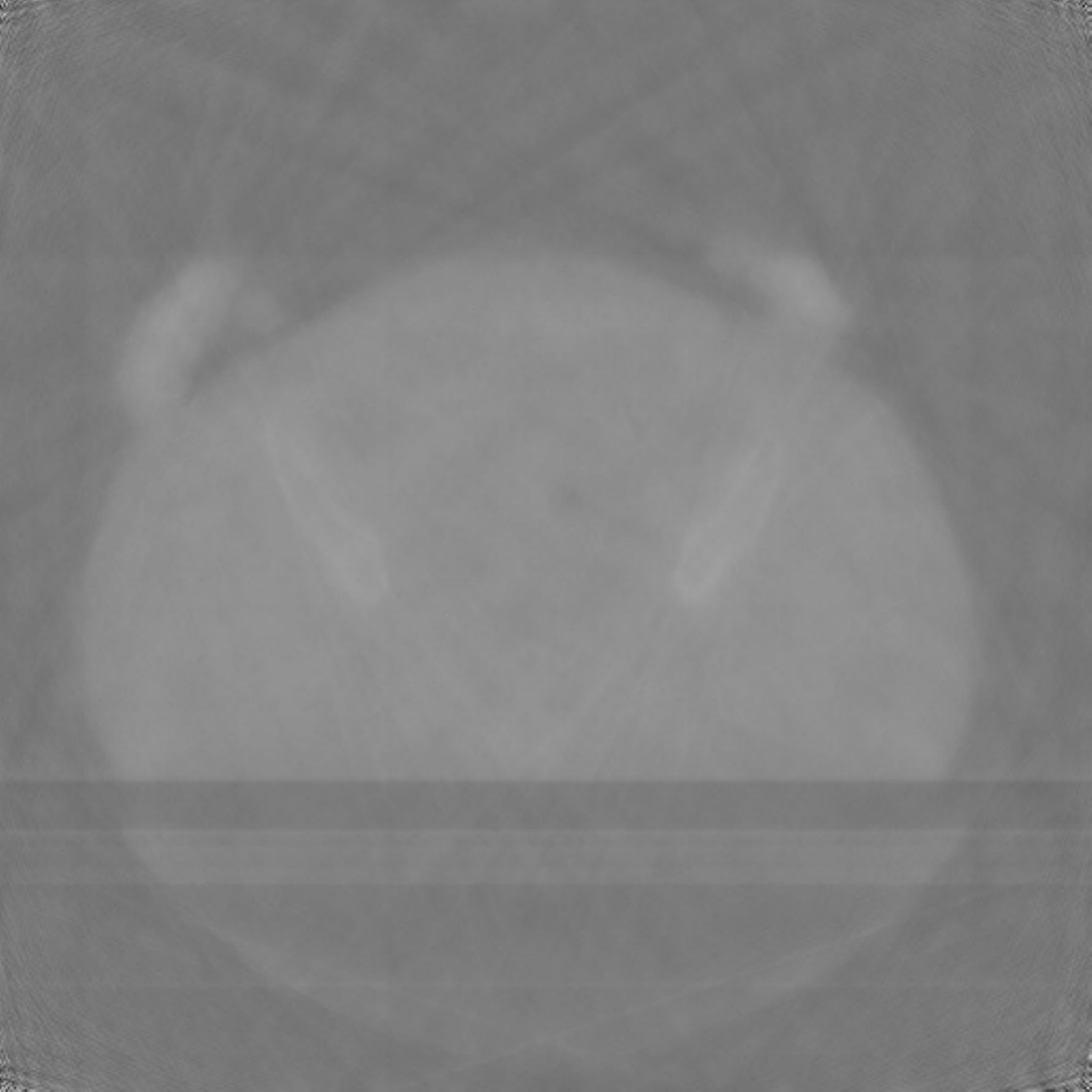}
        \caption{}
    \end{subfigure}
\par\bigskip 
    \begin{subfigure}[b]{0.45\linewidth}
        \includegraphics[width=\textwidth]{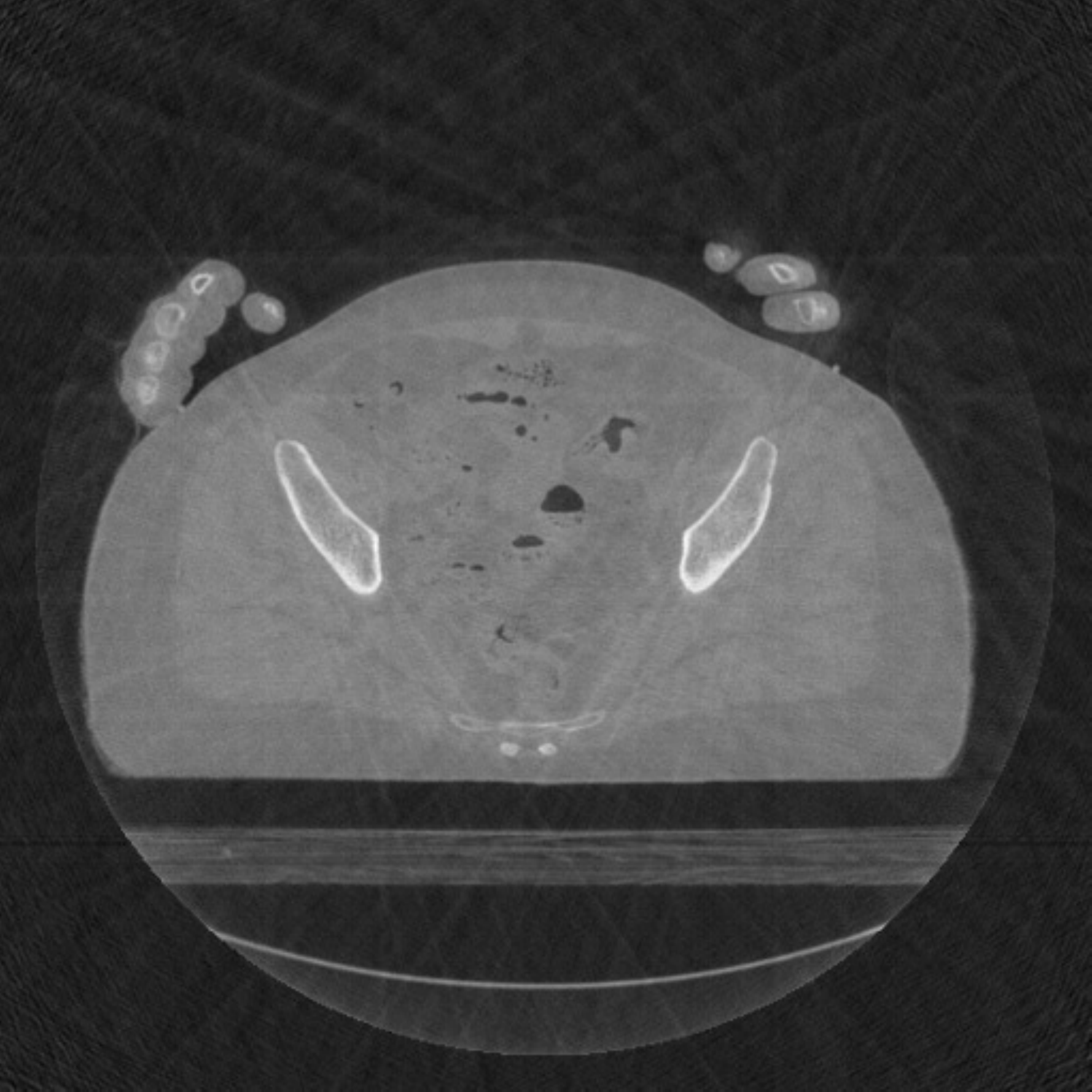}
        \caption{}
     \end{subfigure}
    \begin{subfigure}[b]{0.45\linewidth}
        \includegraphics[width=\textwidth]{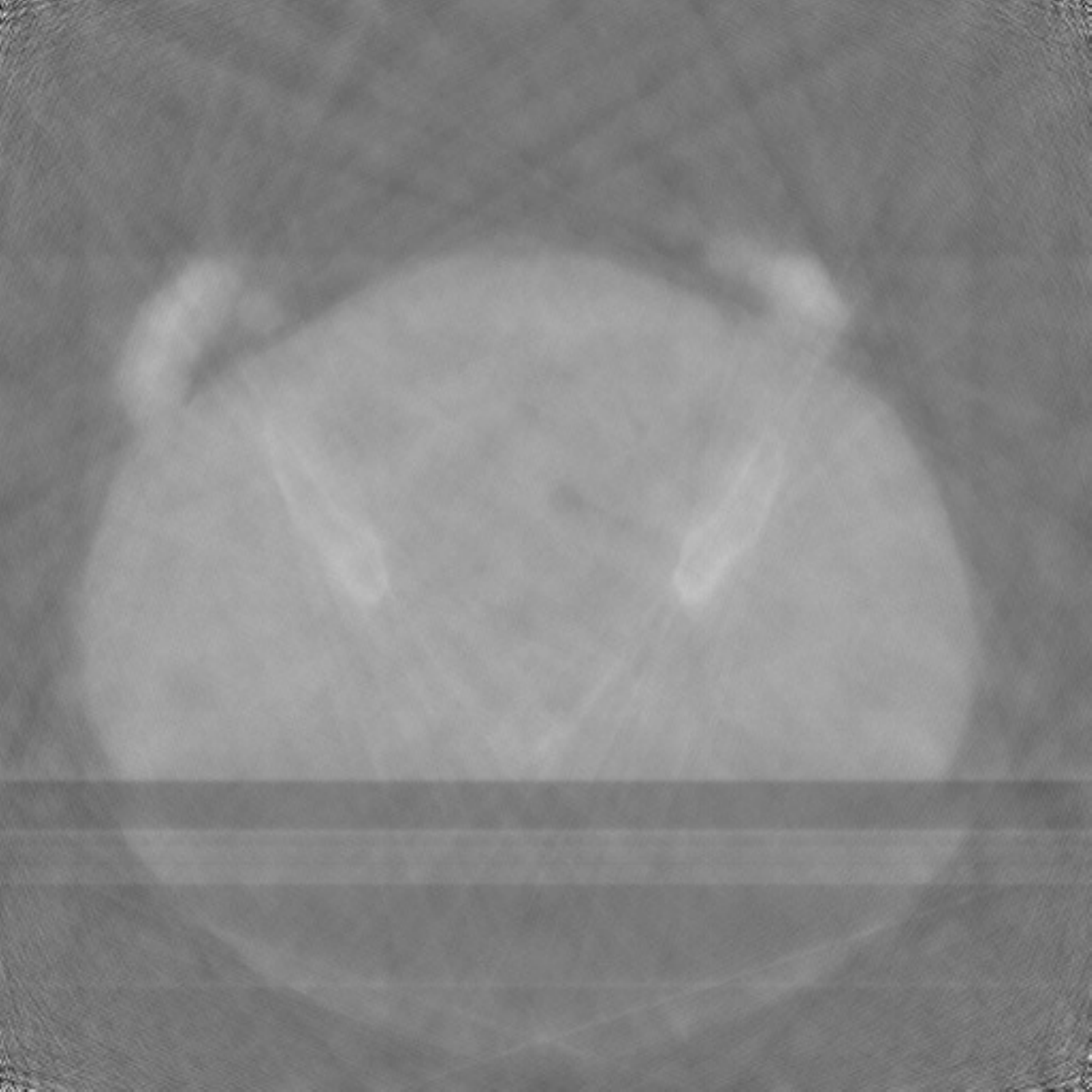}
        \caption{}
     \end{subfigure}
    \caption{Reconstruction of a slice from the pelvis CT dataset~\cite{pelvis} from measurements along 18 angles and with 2\% noise. (a) Reconstructed by filtered backprojection (b) Reconstructed slice using plain CS, without any prior (c) Reconstructed slice using CS, with global prior (d) Reconstructed slice using CS, with patch based prior with patch size = $8\times 8$ pixels.}
\label{fig:global_local_pelvis}
\end{figure}

\begin{figure}[!h]
\centering
\begin{subfigure}[b]{0.85\linewidth}
        \includegraphics[width=\linewidth]{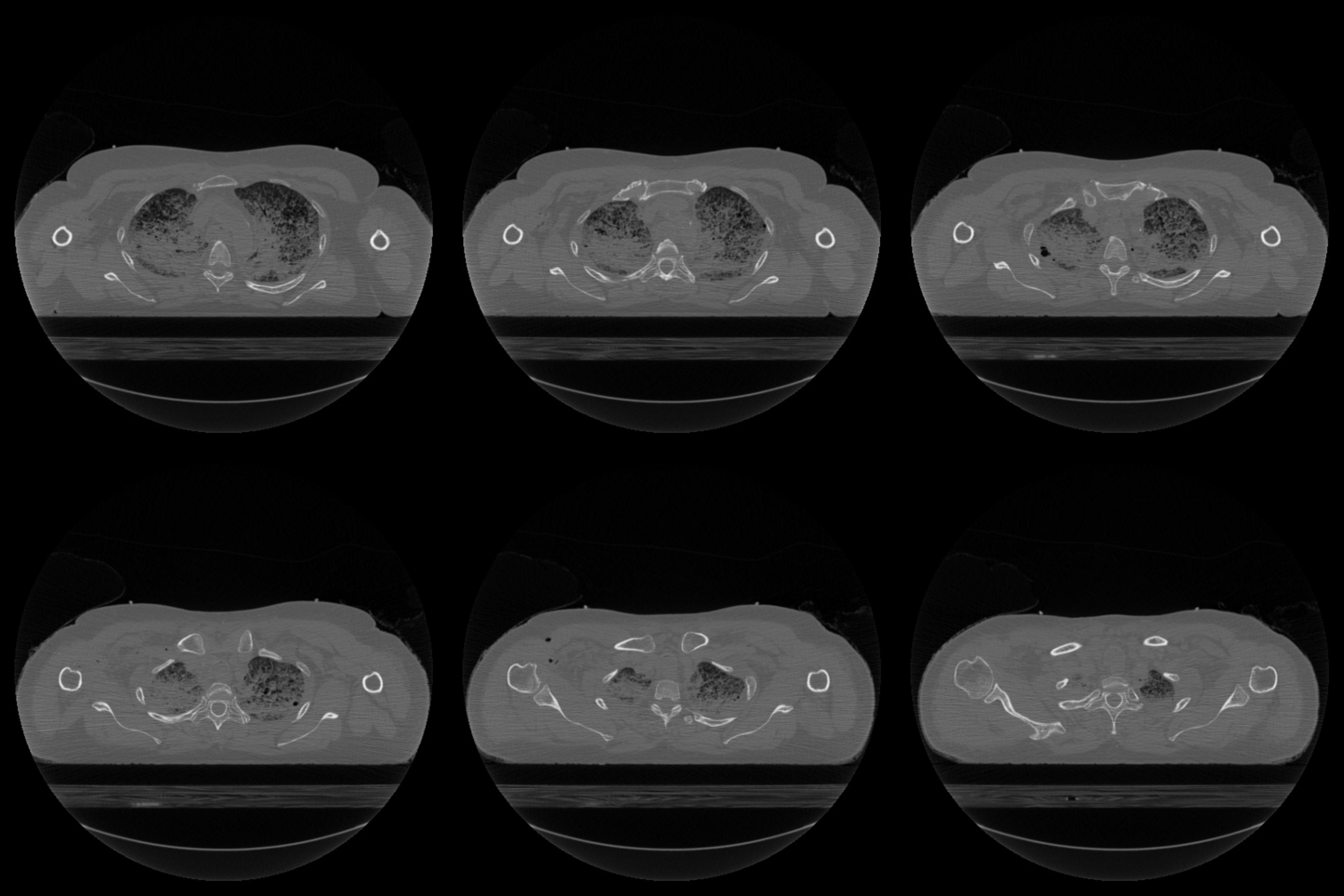}
\caption{}
\label{fig:shoulder_templates}
\end{subfigure}
    \begin{subfigure}[b]{0.3\linewidth}
        \includegraphics[width=\textwidth]{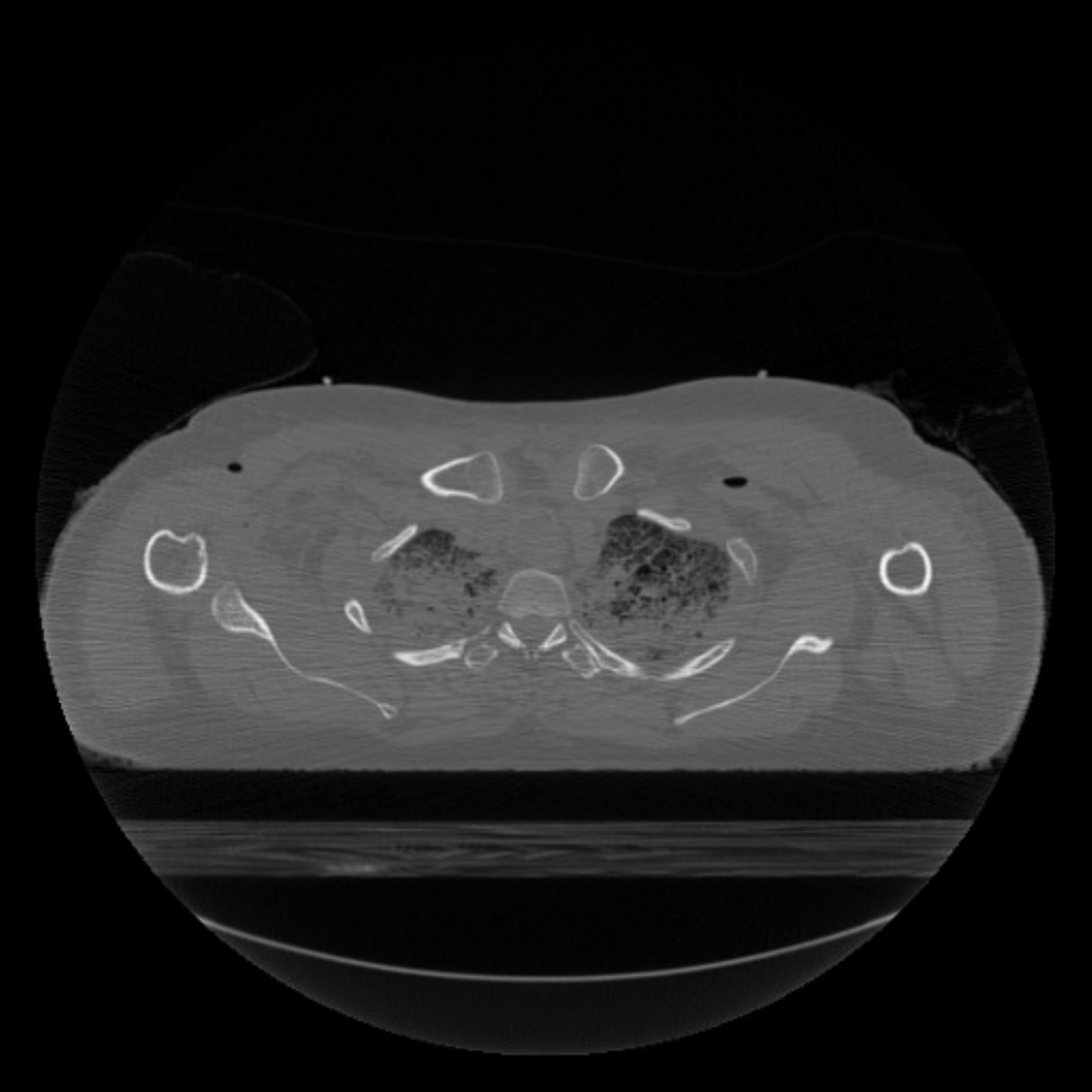}
        \caption{}
\label{fig:shoulder_test}
     \end{subfigure}
        \caption{Data from shoulder CT dataset~\cite{shoulder}. Each slice is of size $512\times 512$. (a) Six templates (slice nos. 350, 360, 370, 380, 390 and 400 from the CT volume) that were used to build the prior. (b) The test data (slice no. 385 from the same volume).}
\end{figure}

\begin{figure}[!h]
\centering
    \begin{subfigure}[b]{0.4\linewidth}
        \includegraphics[width=\textwidth]{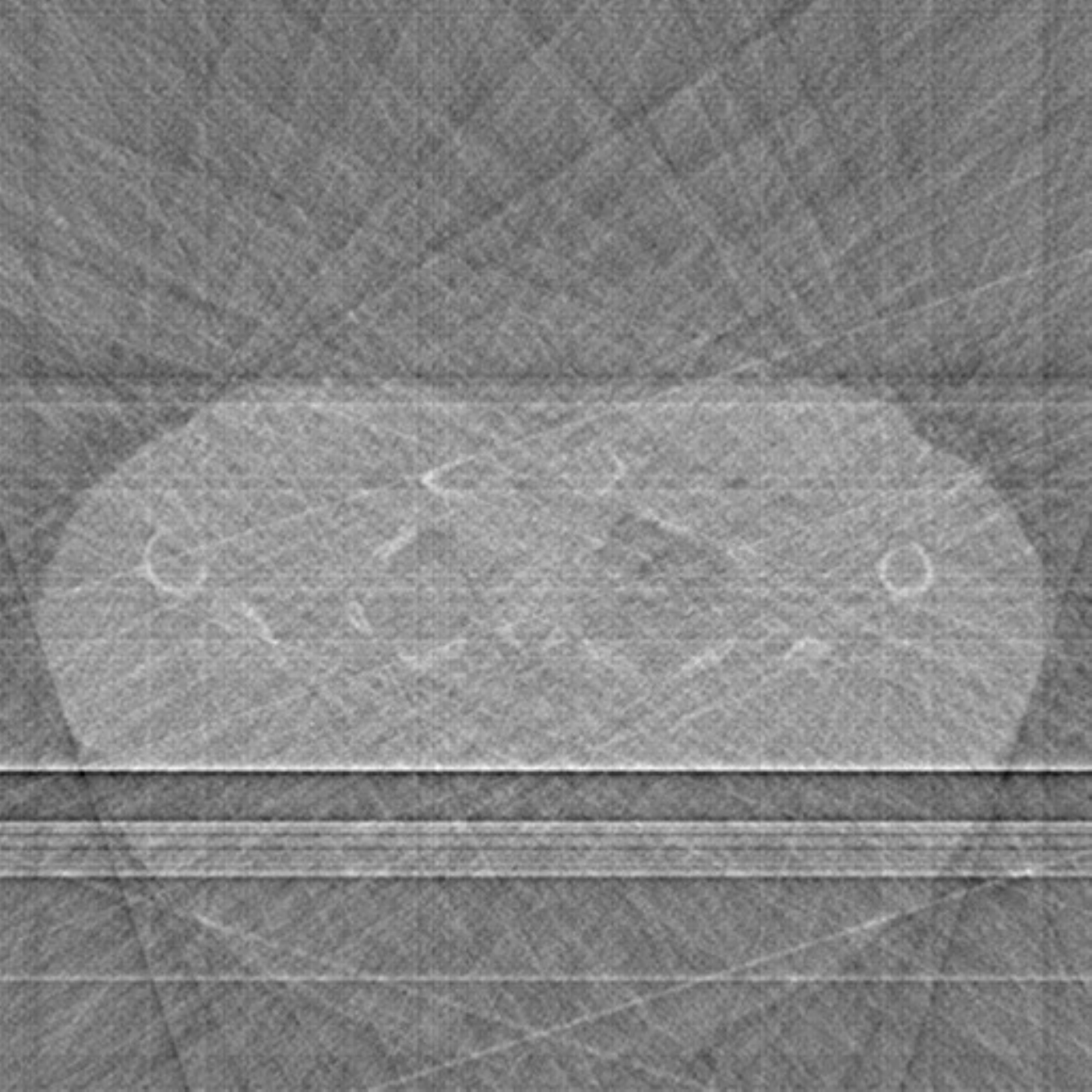}
        \caption{}
     \end{subfigure}
    \begin{subfigure}[b]{0.4\linewidth}
        \includegraphics[width=\textwidth]{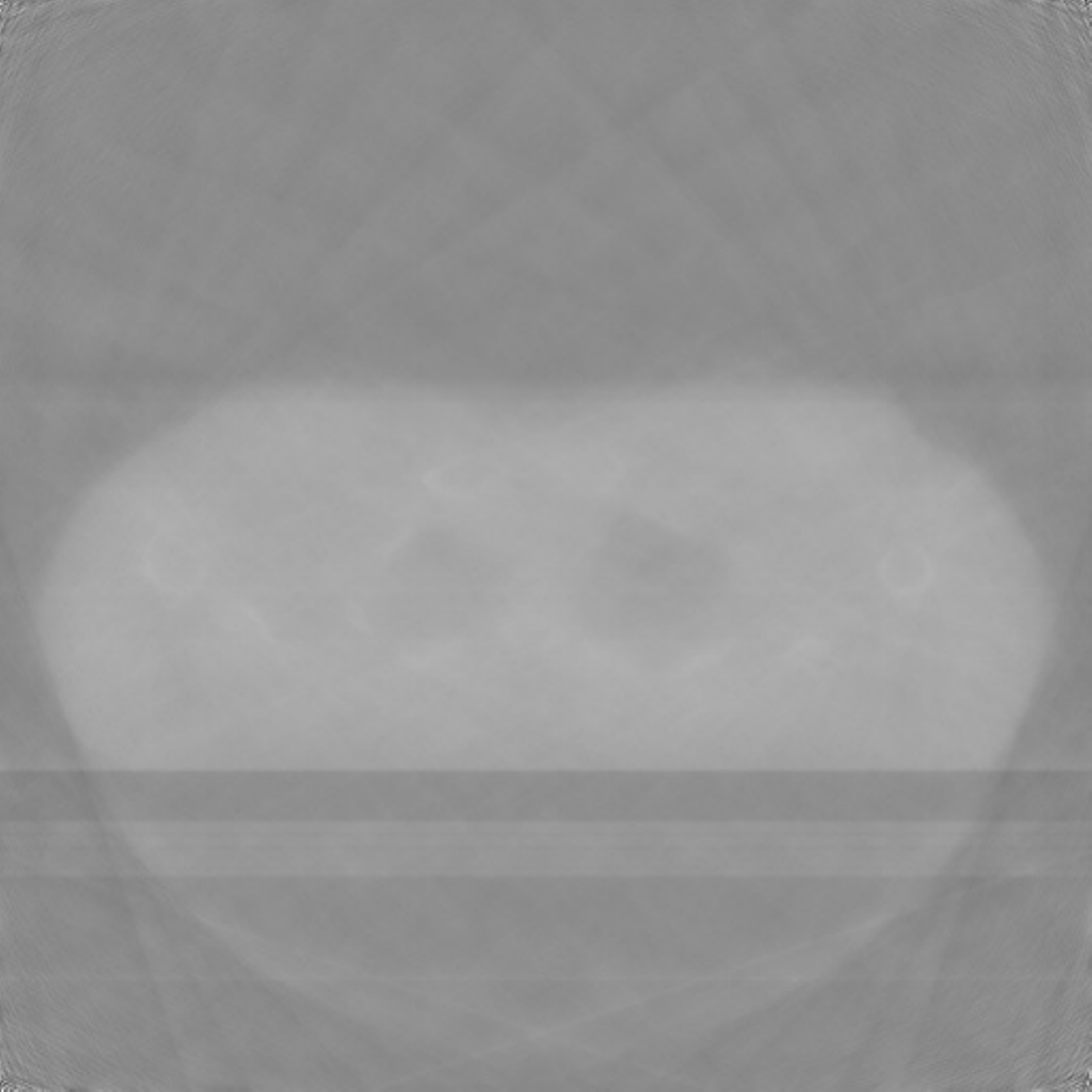}
        \caption{}
    \end{subfigure}
\par\bigskip 
    \begin{subfigure}[b]{0.4\linewidth}
        \includegraphics[width=\textwidth]{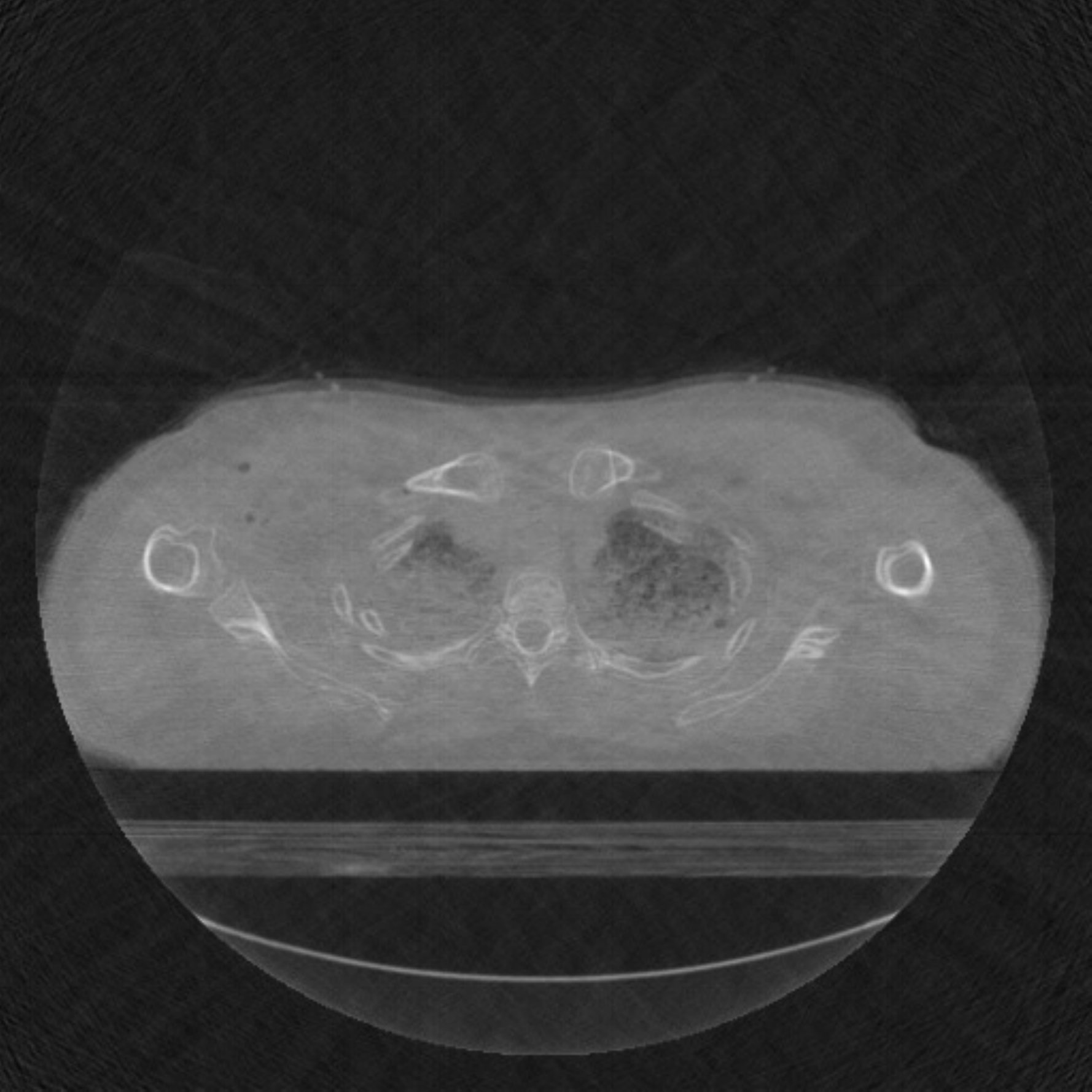}
        \caption{}
     \end{subfigure}
    \begin{subfigure}[b]{0.4\linewidth}
        \includegraphics[width=\textwidth]{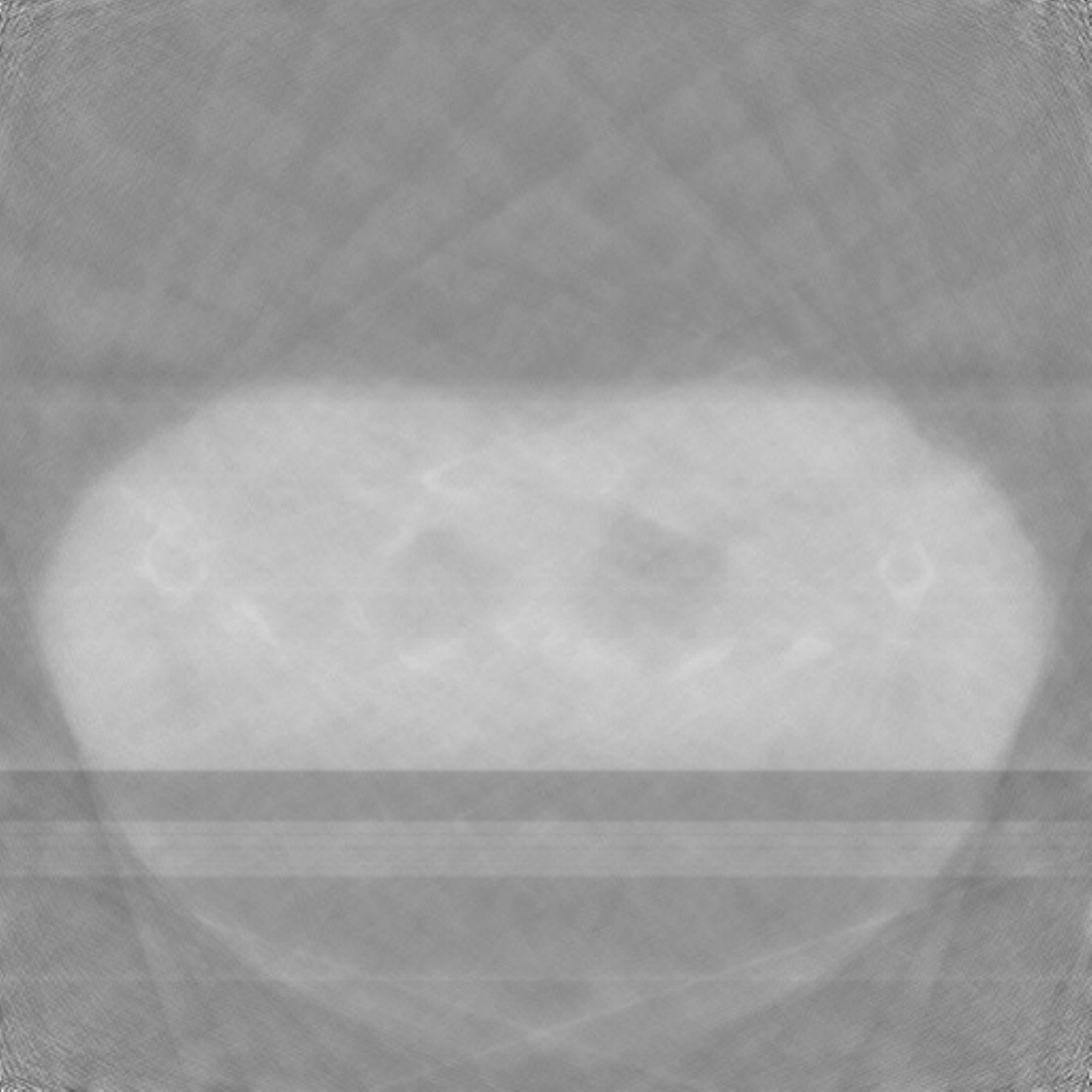}
        \caption{}
     \end{subfigure}
    \caption{Reconstruction of a slice from the shoulder CT dataset~\cite{shoulder} from measurements along 16 angles and with 2\% noise. (a) Reconstructed by filtered backprojection (b) Reconstructed slice using plain CS, without any prior (c) Reconstructed slice using CS, with global prior (d)  Reconstructed slice using CS, with patch based prior with patch size = $8\times 8$ pixels.}
\label{fig:global_local_shoulder}
\end{figure}
\section{Comparison with Patch based techniques}
\label{sec:patch_based}
We have compared the performance of the proposed PCA based global prior with the patch based dictionary prior, both using CS recovery routines. To build the patch based dictionary, we use the K-SVD algorithm. The template slices are divided into patches and an overcomplete dictionary is learned using the extracted patches. The K-SVD algorithm aims to sparsely represent all the patches using the dictionary. Once this dictionary is computed offline, the reconstruction of a new test slice is performed by minimizing the following objective function:

\begin{multline}
E(\mathbf{\theta},\mathbf{\alpha}) = ||\Phi\Psi\mathbf{\theta}- y||_2^2  + \lambda_1||\mathbf{\theta}||_1 + \frac{\lambda_2}{N_p}\sum_{i=1}^{N_p}||P_i\Psi\mathbf{\theta}-D\alpha_i||_2^2 \\+\frac{\lambda_3}{ N_p}\sum_{i=1}^{N_p}||\mathbf{\alpha_i}||_1
\label{Eq:main_patchBased}
\end{multline}

 where D is the dictionary learned, $N_p$ is the total number of patches and $P_i$ is the operator to extract the $i^{th}$ patch from the intermediate reconstructed slice. The reconstruction is performed by alternate minimization of Eq~\ref{Eq:primaryObj_patchBased} and Eq.~\ref{Eq:alpha_patchBased}, both using  $l1\_ls$ solver.   

\begin{equation}
E_{\mathbf{\alpha}}(\mathbf{\theta}) = ||\Phi\Psi\mathbf{\theta}- y||_2^2  + \lambda_1||\mathbf{\theta}||_1 + \frac{\lambda_2}{N_p}\sum_{i=1}^{N_p}||P_i\Psi\mathbf{\theta}-D\alpha_i||_2^2 
\label{Eq:primaryObj_patchBased}
\end{equation}
For each $i = 1 ..N_p$,
\begin{equation}
E_{\mathbf{\theta}}(\mathbf{\alpha}_i) =  ||P_i\Psi\mathbf{\theta}-D\alpha_i||_2^2  + \frac{\lambda_3}{\lambda_2}||\mathbf{\alpha_i}||_1
\label{Eq:alpha_patchBased}
\end{equation}

This method requires prior tuning of three parameters: $\lambda_1,\lambda_2$ and $\lambda_3$. This is done by reconstructing one of the template slices \textit{a priori}.
\section{Experiments and Results}
\label{sec:experiments}

 We have built the template set using 5-7 slices from a volume of each dataset.
Using more slices may not necessarily help if a subset is sufficient to represent all the variations that the test slice could possess. \emph{The chosen test slices are similar to, but not identical to any of the template slices}. However, in our experiments so far, the templates and test slices have been chosen from the same volume. Figures~\ref{fig:walnut_templates},~\ref{fig:colon_templates},~\ref{fig:humerus_templates},~\ref{fig:pelvis_templates} and~\ref{fig:shoulder_templates} show the chosen template sets for each of the five datasets and figures~\ref{fig:walnut_test},~\ref{fig:colon_test},~\ref{fig:humerus_test},~\ref{fig:pelvis_test} and~\ref{fig:shoulder_test} show the corresponding test slices. To build the patch based dictionary, we have chosen patch size of $8\times 8$ for the pelvis, colon, shoulder and humerus datasets and patch size of $13\times 13$ for walnut dataset. In all the datasets, iid Gaussian noise with standard deviation equal to $2\%$ of mean of the projections was added to the measurements before reconstruction.

The global PCA based method has been compared with the following methods: a) filtered backprojection (using `Cosine' filter), b) plain CS technique that uses only sparsity of DCT prior (i.e. with $\lambda_2$ and $\lambda_3$ set to 0); and c) plain CS DCT prior + patch based dictionary. The visual reconstruction results  are shown in figures~\ref{fig:global_local_walnut},~\ref{fig:global_local_colon},~\ref{fig:global_local_humerus},~\ref{fig:global_local_pelvis} and~\ref{fig:global_local_shoulder}, and the relative mse and the Structural SIMilarity (ssim) index of the results are shown in tables~\ref{tab:walnut},~\ref{tab:colon},~\ref{tab:humerus},~\ref{tab:pelvis} and ~\ref{tab:shoulder}.

Relative mse refers to the relative mean squared error and is given by $(1/n)(\sum_i (y_i - x_i)^2/\sum_i(x_i)^2)$ where $n$ refers to the number of pixels. SSIM~\cite{ssim} is a better evaluation metric that computes the similarity in shape and structure of the contents of one image to that of the other.

It is evident that the PCA based prior (coupled within the CS framework) works best. The patch based dictionary doesn't contribute much to the further improvement over plain CS technique. In previous work, the performance of the patch based dictionaries has not been compared with plain CS reconstruction, as has been done here.

All experiments were performed on MATLAB on a 14-core Xeon CPU with a processor at 2.6GHz and RAM of 32GB. Table~\ref{tab:running_time} shows the approximate time taken for training and reconstruction phases for experiments on a $260\times260$ sized slice from measurements along 12 angles and with $2\%$ noise. Computationally, the global prior based method is much faster than the patch based dictionary method during both the offline prior generation process and the online reconstruction process.

\begin{table}[!h]
\centering
\setlength\extrarowheight{3pt}
\caption{Performance of reconstruction in walnut dataset.}
\label{tab:walnut}
\begin{tabular}{|l|l|l|}
\hline
\textbf{Method} & \textbf{relative mse} & \textbf{ssim} \\ \hline
Filtered  backprojection & 0.4580 & 0.1447 \\ \hline
Plain Compressive Sensing & 0.7724 & 0.0230 \\ \hline
Patch based dictionary prior & 0.8511 & 0.0103 \\ \hline
Global PCA based prior & 0.0172 & 0.5150 \\ \hline
\end{tabular}
\end{table}

\begin{table}[!h]
\centering
\setlength\extrarowheight{3pt}
\caption{Performance of reconstruction in colon dataset.}
\label{tab:colon}
\begin{tabular}{|l|l|l|}
\hline
\textbf{Method} & \textbf{relative mse} & \textbf{ssim} \\ \hline
Filtered  backprojection & 0.5844 & 0.0505 \\ \hline
Plain Compressive Sensing & 0.5266 & 0.0261 \\ \hline
Patch based dictionary prior & 0.7951 & 0.0192 \\ \hline
Global PCA based prior & 0.0280 & 0.1783 \\ \hline
\end{tabular}
\end{table}

\begin{table}[!h]
\centering
\setlength\extrarowheight{4pt}
\caption{Performance of reconstruction in humerus dataset.}
\label{tab:humerus}
\begin{tabular}{|l|l|l|}
\hline
\textbf{Method} & \textbf{relative mse} & \textbf{ssim} \\ \hline
Filtered  backprojection & 1.4089 & 0.0669 \\ \hline
Plain Compressive Sensing & 0.2969 & 0.0743 \\ \hline
Patch based dictionary prior & 0.2839 & 0.0777 \\ \hline
Global PCA based prior & 0.2718 & 0.0759 \\ \hline
\end{tabular}
\end{table}

\begin{table}[!h]
\centering
\setlength\extrarowheight{4pt}
\caption{Performance of reconstruction in pelvis dataset.}
\label{tab:pelvis}
\begin{tabular}{|l|l|l|}
\hline
\textbf{Method} & \textbf{relative mse} & \textbf{ssim} \\ \hline
Filtered  backprojection & 0.4081 & 0.0418 \\ \hline
Plain Compressive Sensing & 0.3131 & 0.0438 \\ \hline
Patch based dictionary prior & 0.0975 & 0.0656 \\ \hline
Global PCA based prior & 0.0274 & 0.1366 \\ \hline
\end{tabular}
\end{table}

\begin{table}[!h]
\centering
\setlength\extrarowheight{4pt}
\caption{Performance of reconstruction in shoulder dataset.}
\label{tab:shoulder}
\begin{tabular}{|l|l|l|}
\hline
\textbf{Method} & \textbf{relative mse} & \textbf{ssim} \\ \hline
Filtered  backprojection & 0.4801 & 0.0387 \\ \hline
Plain Compressive Sensing & 0.3333 & 0.0428 \\ \hline
Patch based dictionary prior & 0.1154 & 0.0627 \\ \hline
Global PCA based prior & 0.0219 & 0.1389 \\ \hline
\end{tabular}
\end{table}

\begin{table}[]
\centering
\setlength\extrarowheight{4pt}
\caption{Running time comparison}
\begin{tabular}{|l|l|l|}
\hline
\textbf{} & \textbf{Global PCA based prior} & \textbf{Patch based dictionary} \\ \hline
Training & 5 sec. & 3 hrs. \\ \hline
Testing & 5 min. & 20 min. \\ \hline
\end{tabular}

\label{tab:running_time}
\end{table}

\section{Conclusion and Future Work}
\label{sec:happy_ending}
With an aim to reconstruct data from exceedingly sparse measurements,
we propose to use a PCA based global prior within a Compressed Sensing
framework. We have validated our method on 2D slices of multiple,
diverse datasets. When sparse measurements of $5\%-10\%$ of the full
Nyquist sampling are used, our results are significantly better than
the state of the art. In the future, the proposed method needs to be
extended for reconstruction of test slice from different subject
volumes, and for the case of three dimensional data sets.

\bibliographystyle{IEEEbib}
\bibliography{prior_ref.bib}

\begin{thebibliography}{10}

\bibitem{Donoho}
D.L. Donoho,
\newblock ``Compressed sensing,''
\newblock {\em IEEE Transactions on Information Theory}, vol. 52, no. 4, pp.
  1289--1306, April 2006.

\bibitem{Preeti2016}
Preeti Gopal, Sharat Chandran, Imants Svalbe, and Ajit Rajwade,
\newblock ``Multi-slice tomographic reconstruction: To couple or not to
  couple,''
\newblock in {\em Proceedings of the Tenth Indian Conference on Computer
  Vision, Graphics and Image Processing}, New York, NY, USA, 2016, ICVGIP '16,
  pp. 85:1--85:7, ACM.

\bibitem{PICCS}
Guang-Hong Chen, Jie Tang, and Shuai Leng,
\newblock ``Prior image constrained compressed sensing {(PICCS)}: A method to
  accurately reconstruct dynamic {CT} images from highly undersampled
  projection data sets,''
\newblock {\em Medical Physics}, vol. 35, no. 2, pp. 660--663, 2008.

\bibitem{liu2016}
J.~Liu, Y.~Hu, J.~Yang, Y.~Chen, H.~Shu, L.~Luo, Q.~Feng, Z.~Gui, and
  G.~Coatrieux,
\newblock ``3{D} feature constrained reconstruction for low dose {CT}
  imaging,''
\newblock {\em IEEE Transactions on Circuits and Systems for Video Technology},
  vol. PP, no. 99, pp. 1--1, 2016.

\bibitem{Xu2012}
Q.~Xu, H.~Yu, X.~Mou, L.~Zhang, J.~Hsieh, and G.~Wang,
\newblock ``Low-dose {X}-ray {CT} reconstruction via dictionary learning,''
\newblock {\em IEEE Transactions on Medical Imaging}, vol. 31, no. 9, pp.
  1682--1697, Sept 2012.

\bibitem{song2014}
Ying Song, Zhen Zhu, Yang Lu, Qiegen Liu, and Jun Zhao,
\newblock ``Reconstruction of magnetic resonance imaging by three-dimensional
  dual-dictionary learning,''
\newblock {\em Magnetic Resonance in Medicine}, vol. 71, no. 3, pp. 1285--1298,
  2014.

\bibitem{Li2016}
Jiansen Li, Ying Song, Zhen Zhu, and Jun Zhao,
\newblock ``Highly undersampled {MR} image reconstruction using an improved
  dual-dictionary learning method with self-adaptive dictionaries,''
\newblock {\em Medical {\&} Biological Engineering {\&} Computing}, pp. 1--16,
  2016.

\bibitem{Lior2015}
Lior Weizman, Yonina~C. Eldar, and Dafna Ben~Bashat,
\newblock ``Compressed sensing for longitudinal {MRI}: An adaptive-weighted
  approach,''
\newblock {\em Medical Physics}, vol. 42, no. 9, pp. 5195--5208, 2015.

\bibitem{candes}
Emmanuel~J Cand{\`e}s, Justin Romberg, and Terence Tao,
\newblock ``Robust uncertainty principles: Exact signal reconstruction from
  highly incomplete frequency information,''
\newblock {\em Information Theory, IEEE Transactions on}, vol. 52, no. 2, pp.
  489--509, 2006.

\bibitem{Luong_2016}
H.~Van Luong, J.~Seiler, A.~Kaup, and S.~Forchhammer,
\newblock ``Sparse signal reconstruction with multiple side information using
  adaptive weights for multiview sources,''
\newblock in {\em 2016 IEEE International Conference on Image Processing
  (ICIP)}, Sept 2016, pp. 2534--2538.

\bibitem{Zhan2016}
Z.~Zhan, J.~F. Cai, D.~Guo, Y.~Liu, Z.~Chen, and X.~Qu,
\newblock ``Fast multiclass dictionaries learning with geometrical directions
  in {MRI} reconstruction,''
\newblock {\em IEEE Transactions on Biomedical Engineering}, vol. 63, no. 9,
  pp. 1850--1861, Sept 2016.

\bibitem{walnut}
``{Walnut {CT} dataset},'' \url{https://www.uni- muenster.de/Voreen/}, last
  viewed--April, 2017.

\bibitem{l1ls}
K.~Koh, S.-J. Kim, and S.~Boyd,
\newblock ``l1-ls: Simple matlab solver for l1-regularized least squares
  problems,'' \url{https://stanford.edu/~boyd/l1_ls/}, last viewed--July, 2016.

\bibitem{colon}
``{CT Colonography},'' \url{https://idash.ucsd.edu/data-collections}, last
  viewed--April, 2017.

\bibitem{humerus}
``{Humerus {CT} dataset},'' \url{http://isbweb.org/data/vsj/humeral/}, last
  viewed--July, 2016.

\bibitem{pelvis}
``{Pelvis {CT} dataset},'' \url{https://medicine.uiowa.edu/}, last
  viewed--June, 2017.

\bibitem{shoulder}
``{Shoulder {CT} dataset},'' \url{https://medicine.uiowa.edu/}, last
  viewed--June, 2017.

\bibitem{ssim}
Zhou Wang, A.~C. Bovik, H.~R. Sheikh, and E.~P. Simoncelli,
\newblock ``Image quality assessment: from error visibility to structural
  similarity,''
\newblock {\em IEEE Transactions on Image Processing}, vol. 13, no. 4, pp.
  600--612, April 2004.

\end{thebibliography}

\end{document}